\documentclass[pdflatex, sn-mathphys]{sn-jnl}

\jyear{2023}%

\theoremstyle{thmstyleone}%
\theoremstyle{thmstyletwo}%

\theoremstyle{thmstylethree}%

\def\bng{\bngx}

\font\bngx=bang10

\def\*#1*#2{o\null{#2}{#1}}

\def\sh#1{\setbox0=\hbox{#1}%
     \kern-.02em\copy0\kern-\wd0
     \kern.04em\copy0\kern-\wd0
     \kern-.02em\raise.0433em\box0 }
\usepackage[T1]{fontenc}
\usepackage{amsmath,amssymb,amsfonts}
\usepackage{hyperref}
\usepackage{array}
\usepackage{enumerate} 
\usepackage{subcaption}
\usepackage{graphicx}
\usepackage{multirow}
\usepackage{longtable}
\usepackage{pdflscape}

\newcommand{\hochkomma}{$^{,}$}
\usepackage{xurl}
\usepackage{float}
\usepackage{graphicx}
\usepackage{caption}
\usepackage{subcaption}
\usepackage{enumitem}

\usepackage{etoolbox}
\makeatletter
\patchcmd{\ps@headings}
{\hbox to \hsize{\hfill Springer Nature 2021 \LaTeX\ template\hfill}}
{\hbox to \hsize{}}
{}
{}
\patchcmd{\ps@headings}
{\hbox to \hsize{\hfill Springer Nature 2021 \LaTeX\ template\hfill}}
{\hbox to \hsize{}}
{}
{}
\patchcmd{\ps@titlepage}
{\hbox to \hsize{\hfill Springer Nature 2021 \LaTeX\ template\hfill}}
{\hbox to \hsize{}}
{}
{}
\makeatother

\begin{document}

\title[Tackling Fake News in Bengali]{Tackling Fake News in Bengali: Unraveling the Impact of Summarization vs. Augmentation on Pre-trained Language Models}


\author*[]{\fnm{Arman Sakif} \sur{Chowdhury}}\email{armansakif90@gmail.com}
\equalcont{These authors contributed equally to this work.}

\author{\fnm{G. M.} \sur{Shahariar}}\email{sshibli745@gmail.com}
\equalcont{These authors contributed equally to this work.}

\author{\fnm{Ahammed Tarik} \sur{Aziz}}\email{tarikaziz821@gmail.com}

\author{\fnm{Syed Mohibul} \sur{Alam}}\email{sadeemohib@gmail.com}

\author{\fnm{Md. Azad} \sur{Sheikh}}\email{mazad6932@gmail.com}

\author{\fnm{Tanveer Ahmed} \sur{Belal}}\email{belal.cse@aust.edu}

\affil{\orgname{Ahsanullah University of Science and Technology}, \orgaddress{\city{Dhaka}, \country{Bangladesh}}}


\abstract{With the rise of social media and online news sources, fake news has become a significant issue globally. However, the detection of fake news in low resource languages like Bengali has received limited attention in research. In this paper, we propose a methodology consisting of four distinct approaches to classify fake news articles in Bengali using summarization and augmentation techniques with five pre-trained language models. Our approach includes translating English news articles and using augmentation techniques to curb the deficit of fake news articles. Our research also focused on summarizing the news to tackle the token length limitation of BERT based models. Through extensive experimentation and rigorous evaluation, we show the effectiveness of summarization and augmentation in the case of Bengali fake news detection. We evaluated our models using three separate test datasets. The BanglaBERT Base model, when combined with augmentation techniques, achieved an impressive accuracy of 96\% on the first test dataset. On the second test dataset, the BanglaBERT model, trained with summarized augmented news articles achieved 97\% accuracy. Lastly, the mBERT Base model achieved an accuracy of 86\% on the third test dataset which was reserved for generalization performance evaluation. The datasets and implementations are available at \url{https://github.com/arman-sakif/Bengali-Fake-News-Detection}. 
}

\keywords{fake news, classification, summarization, Bengali, augmentation, transformers, BERT}



\maketitle

\section{Introduction}\label{sec1}
Fake news refers to news stories or articles that are deliberately misleading or fabricated and spread through various media channels, such as social media, news websites, or even traditional printed newspapers \cite{bib13}. It can also be a misguided viewpoint of the author. ``Fake news'' is an old term, dating back to the time when newspapers had just gained popularity, long before the web was invented \cite{bib14}. These types of news articles are intentionally created to deceive readers or viewers or to manipulate public opinion for political, financial, or other reasons. Fake news is created and spread by either those who have ideological interests, malicious intentions, or media individuals looking for some views, which are commonly termed as clickbait. Fake news can be presented in a variety of formats, such as articles, images, videos, and even memes \cite{bib15}.

The dissemination of fake news has led to significant societal consequences, including polarization, misinformation, and confusion. A study \cite{bib1} shows that fake news spreads faster than the truth and humans are primarily responsible for the spread of misleading information. Due to the fact that most of these stories are not verified and are so much more interesting than regular news, it makes it difficult to resist spreading them. According to a 2018 study by Pew Research Center, 64\% of U.S. adults believe that fake news has caused a great deal of confusion about the basic facts of current events, and 23\% of people say they have shared a made-up news story \cite{bib2}.

On February 3, 2022, a prominent Bangladeshi TV station had a news story rated as false information by the fact-checking unit of the international news agency AFP \cite{bib16}. The story, which was credited to the state-owned Saudi Press Agency (SPA), claimed that the Saudi government had approved a draft amendment to redesign its national flag by removing Kalima Tayyiba (Islamic declaration of faith) from it. However, this claim turned out to be untrue as neither the SPA nor any other Saudi news media had reported anything of that sort. The actual revision of the draft was related to proposing new regulations regarding the use of the flag, and several Saudi newspapers confirmed that no changes were proposed to the contents of the flag \cite{bib17}. In a separate report by Boom Bangladesh \cite{bib18}, a third-party fact-checking partner of Facebook focusing on Bangladesh, it was revealed that they had detected 23 instances of fake news in mainstream media outlets between March and December 2020. Notably, the most popular TV channel in the country was caught spreading fake news 10 times during that 10-month period \cite{bib17}.

Fake news can also have serious consequences for public health. During the early days of the Coronavirus period in 2020, a rumor spread out that drinking disinfectants could cure Covid-19. Some people even ingested hand sanitizer with methanol which led 4 people to death and 26 hospitalized \cite{bib5}. In another instance, people believed that 5G spread coronavirus, and more than 70 phone masts have been vandalized because of this false rumor in the UK \cite{bib6}. Some people even claimed the whole coronavirus is a hoax and refused to take quarantine seriously or taking cure when it came out. 
Fake news can also impact businesses and their reputations. For example, false rumors about a company's financial situation can lead to a decline in stock prices and cause significant financial losses. As we can see, addressing the challenge of identifying and classifying fake news has become increasingly important in the context of the digital age. 

Research supports that fake news are harder to detect \cite{bib7}. There are various reasons for fake news being difficult to detect. With the proliferation of social media platforms and the internet, there are now millions of sources of news and information. Many of these sources are not reputable, and it can be difficult to distinguish between legitimate sources and fake news. As already discussed, fake news spreads even faster than real news. By the time a news article has been identified as fake, it may have already been shared by thousands or even millions of people. With the sheer number of people sharing the same news, people tend to believe the news to be real. Another reason can be confirmation bias \cite{bib8}. People often seek out information that already confirms their existing beliefs and opinions. This means that fake news stories that confirm people's biases are more likely to be shared and believed. 

In recent years, machine learning has evolved as an appropriate solution for fake news detection for several reasons. Firstly, for a machine learning (ML) model to work, it needs to learn with a significant amount of data. And with the proliferation of social media and online news sources, there is now a significant amount of data available to train machine learning models to detect fake news. Machine learning algorithms process large amounts of data automatically and find underlying patterns or characteristics which can help detect fake news. Pattern recognition is particularly useful as many fake news share an underlying pattern. Research in fake news has been conducted for a long time, especially in English. There has been extensive work on classifying fake news in the English language \cite{bib20,bib21,bib22,bib23,bib24,bib25,bib26}. Recently other than the English language, research on fake news is being conducted in Spanish \cite{bib27}, French \cite{bib28,bib29,bib30,bib31}, Arabic \cite{bib32}, Chinese \cite{bib33,bib34,bib35,bib36,bib37,bib38,bib39}, and many other languages \cite{bib40,bib41,bib42,bib43,bib44,bib45,bib46,bib47,bib48,bib49}. There are many different methods used in these researches but deep learning based methods seemed to be the most popular.

Bengali is one of the largest speaking languages, and although there are English newspapers in our country, most people read news in our native language- Bengali. But there has not been much research to detect fake news in Bengali. The probable reason that there is so little research in Bengali to detect fake news is because there is not enough labeled fake news data available. As there are authentic news sites and most big news organizations have online news portals, it is not difficult to collect authentic news data. But it is difficult to find a source with large amounts of data for fake news. In recent years, the few pieces of research that have been conducted in Bengali mostly used deep learning as a base method to detect fake news, especially using BERT models is prevalent. One of the most notable research is BanFakeNews \cite{bib10}. The limitation of this research is that their built model has trouble detecting the minority class which is fake news. Sharma et al. \cite{bib11} conducted another research to detect satirical news in Bengali using Convoluted Neural Network. In a recent research published in 2022- AugfakeBERT \cite{bib12}, the authors used the BanFakeNews dataset and tried to tackle the problem of insufficient fake news articles through text augmentation. The limitation of this research is that they did not address BERT’s ability to only take input sequences up to 512 tokens. This shortcoming can be significant as many news articles contain large volumes of text. 

In this study, by fine-tuning a few existing pre-trained transformers and employing summarization and augmentation techniques, we have proposed a novel methodology consisting of four distinct approaches for Bengali fake news identification. To tackle the class imbalance problem we increased fake news articles by generating synthetic Bengali fake news through English to Bengali translation and using different types of augmentation such as - token replacements and paraphrasing. To tackle BERT’s 512 token limitation, large texts were summarized using pre-trained summarization models. Additionally, to test the generalization performance of the proposed approaches, we manually collected 102 fake news articles from various sources and selected equal numbers of random authentic news from \textit{BanFakeNews} corpus. We utilized standard evaluation metrics to measure the performance of the proposed methodology. Our experimental results show the ability of the proposed approaches to detect Bengali fake news with high accuracy even with completely unseen test dataset. In summary, the followings are the notable contributions we made in this paper:

\begin{itemize}
\item We propose a novel methodology consisting of four distinct approaches using summarization and augmentation to detect Bengali fake news.
\item Through extensive experimentation, we demonstrate the effectiveness of applying summarization and augmentation methods to enhance the generalization performance. To overcome the token length limitation of BERT-based models, we develop a summarization pipeline. Furthermore, we employ translation and augmentation strategies to address the class imbalance issue resulting from the scarcity of fake news articles in the training dataset.
\item To evaluate the generalization performance, we curate a test dataset comprising 102 manually collected fake news articles that are entirely unseen by the existing models. This dataset serves as a benchmark for comparing the performance of different approaches.
\item We have made the datasets and implementations of the proposed approaches publicly available with the aim of garnering collaboration and encouraging future researchers to contribute to this particular field.
\end{itemize}

To assess the performance of the proposed approaches, we address the following research questions which are discussed in detail in section \ref{sec:Results} of our paper:

\begin{enumerate}[label=(\alph*)]
    \item \textbf{RQ1}: How does fine-tuning without summarization and augmentation perform when translated news (English-Bengali) are combined with Bengali news articles for the task of fake news classification?
    \item \textbf{RQ2}: What impact does introducing summarization before fine-tuning have on fake news detection?
    \item \textbf{RQ3}: How does the utilization of augmented fake data during fine-tuning affect fake news detection?
    \item \textbf{RQ4}: What is the influence of summarizing augmented news articles on fake news classification?
\end{enumerate}

The rest of the paper is organized as follows: Section \ref{sec2} reviews the relevant literature to classify fake news. Section \ref{sec3} illustrates all the corpus we used in our paper. Section \ref{sec4} describes the research methodology. Our experimental results and findings are presented in Section \ref{sec5}. The final Section \ref{sec7} contains a description of the limitations of this work and future possibilities and then draws a conclusion.  

\section{Related Works}\label{sec2}

In this section, we review some of the previous studies related to our research. While extensive research has been conducted on detecting fake news in English, there are only a few notable works in Bengali. We have examined a range of studies in Bengali, English, and several other languages, categorizing them into four main approaches: traditional machine learning, neural networks, hybrid and ensemble methods, and pre-trained transformer models. The following sections provide a detailed discussion of each of these categories.

\subsection{Traditional Approaches}\label{sec2.1}

Using traditional machine learning (ML) based approach, there are several techniques to detect fake news. This approach typically involves feature engineering and the use of various classifiers such as Support Vector Machines (SVM), Logistic Regression (LR), Naive Bayes (NB), Random Forests (RF), K-nearest Neighbor algorithm (KNN), Stochastic Gradient Descent (SGD), Passive Aggressive Classifier (PAC) and Decision Trees (DT).

Khan et al. \cite{bib61} conducted a benchmark study to evaluate the effectiveness of several practical approaches on three distinct English datasets, the largest and most diverse of which was created by them. They experiment with both traditional ML based and neural network based approaches. Within traditional models they used SVM, LR, DT, Multinomial Naive Bayes (MNB) and KNN - and among them Naive Bayes, with n-gram features, performed the best. 

Baarir et al. \cite{bib62} used two existing English datasets and divided news data into three categories: Textual, categorical and numerical. They employed an SVM classifier and experimented with various feature extraction and hyperparameter configurations to enhance accuracy. They identified the most impactful features in the following order: text, author, source, date, and sentiment, ultimately achieved 100\% accuracy when using all features together.
However, such a result is typically considered improbable in real-world classification tasks and may indicate presence of overfitting, where the model memorizes the training data rather than learning generalizable patterns. Their use of source and author features for model training and evaluation may introduce bias. For example, if a dataset predominantly associates certain sources, such as \texttt{100percentfedup.com}, with fake news, the model may learn to classify news based on the source rather than its content, thereby reducing generalizability.

Hossain et al. \cite{bib10} created \textit{BanFakeNews}, which is an annotated dataset of around 48 thousand authentic and around 1300 fake news in Bengali. To our knowledge, this is the only publicly available dataset in on Bengali fake news and it paved research on Bengali fake news detection. They collected authentic news articles from 22 reputed sources and three types of fake news - misleading, clickbait and satire are collected from some popular websites. They initially experiment with traditional ML models - SVM, LR and RF and empirically find that SVM incorporated with all linguistic features outperforms the other two models, achieving 0.91 \( F_1 \)-score on fake class. They also observe that lexical features perform better than other linguistic features. Additionally, they claim that the use of punctuation in fake news is more frequent, and most of the time fake news is found on the least popular sites.

Hussain et al. \cite{bib63} employed MNB and SVM algorithms to construct a Bengali fake news detection model. The model leveraged both count vectorizer and TF-IDF for feature extraction. They created their own dataset of around 2500 news articles among which 1548 were authentic and 993 fake. In their experimentation, SVM with an accuracy of 0.9664 outperforms MNB with an accuracy of 0.9332. While the results are promising, the relatively small size of the dataset raises concerns about the model's ability to generalize effectively to real-world scenarios involving larger and more diverse texts.

Sraboni et al. \cite{bib64} employed the BanFakeNews dataset \cite{bib10} alongside an additional dataset of 2,500 news articles \cite{bib63} for training various traditional ML algorithms, including RF, PAC, MNB, SVM, LR, and DT. TF-IDF served as the feature extraction method. Rather than using all of 51.8k available data, they used 3.5k authentic data and 2.3k fake data to remove bias. They also experimented with different train-test split ratios with 50-50, 60-40, 80-20 and 70-30 and empirically found that 70-30 split ratio gives the best result. Among the models, PAC and SVM achieve 0.938 and 0.935 accuracy respectively which are higher than the other models they trained. However, TF-IDF is a non-contextual method and does not capture the sequential or semantic relationships between words, limiting the model's deeper understanding of the textual content.

Mugdha et al. \cite{bib68} created their own balanced dataset of total 112 instances with an equal number of fake and real news. They tested multiple traditional algorithms with tenfold cross validation - SVM, LR, Multi-layer Perceptron (MLP), RF, Voting Ensemble Classifier (VEC) and Gaussian Naive Bayes (GNB). Initially, none of the algorithms were performing well. Then they applied 3 feature selection techniques - Principal Component Analysis (PCA), Kernel PCA and Extra Tree Classifier (ETC). Through empirical research, they discovered that ETC outperformed other techniques and that GNB with ETC outperformed other models, with an accuracy of 0.8552. While Mugdha et al. explored various traditional ML algorithms and feature selection techniques, their dataset was relatively small, consisting of only 112 instances.

Piya et al.  \cite{bib72} investigated the development of a model capable of concurrent identification of Bengali and English fake news. Their approach for detecting fake news from a bilingual perspective uses feature extraction techniques such as TF-IDF and N-Gram analysis. They trained six supervised traditional ML models - LR, Linear SVC, DT, RF, MNB, and PAC. Among these six models, the linear SVC performed best and score 0.9329 in accuracy and 0.93 \( F_1 \)-score. They adopted BanFakeNews for the Bengali portion of their Dataset. Additionally, they used ISOT, a dataset released in \cite{bib66}, for the English dataset. In their final dataset, they had 42,324 bilingual samples. The majority of them were in English. 

Mugdha et al. \cite{bib73} concentrated on leveraging headlines alone for Bengali fake news detection. They developed their own Bengali dataset with a total of 538 cases, of which 269 were authentic and 269 were false. They have used 9 different traditional ML algorithms for their research - SVM, LR, MLP, RF, VEC, GNB, MNB, AdaBoost (AB) and Gradient Boosting (GB). With a score of 0.87, the GNB classifier outperformed all other models in terms of accuracy. This algorithm chose the attribute using an ETC feature extractor and a text feature dependent on TF-IDF. Performance was assessed using stratified 10-fold cross-validation on the dataset. They contribute to this field by creating a novel dataset and also a new Bengali stemmer. However, the research is limited by the modest size of the training dataset and its singular focus on headline-based analysis.

\subsection{Neural Network based Approaches}\label{sec2.2}
A neural network is a type of computational model that draws inspiration from the structure and operation of the human brain. It is a hierarchy of neurons, or interconnected nodes, arranged in layers. When a neural network contains several layers, it becomes a deep neural network. Deep learning is the usage of neural networks that have numerous hidden layers, which enables them to learn hierarchical data representations. Deep learning approaches, as opposed to traditional machine learning, are able to recognize significant features and grasp the semantic context of textual input. In this section, we discuss the relevant literature which was based on neural network models. 

Sharma et al. \cite{bib11} adopted Convolutional Neural Network (CNN) based approach built upon a hybrid feature extraction model. They created their own dataset by crawling satire news portal - Motikontho for satire news and real news from Prothom ALo and Ittefaq, two popular news media. They gather 1480 satire news and to balance the dataset, they picked the same amount from the real news. For vectorization, They used 1000-dimensional TF-IDF vectorizer and 10-dimensional Word2Vec which resulted in 1000 x 10 x 2 (2D embeddings with separate positive and negative channels) document representation. With this balanced training, they were able to achieve 0.96 in accuracy.

Bahad et al. \cite{bib75} proposed using Bi-directional LSTM recurrent neural network to detect fake news. They used two publicly available news articles datasets separately. They also experimented with CNN, vanilla Recurrent Neural Network (RNN), unidirectional LSTM and found Bi-LSTM to be superior, achieving 0.9108 in accuracy. 

In the paper of Khan et al. \cite{bib61}, along with traditional methods, they also experimented with deep learning models such as - CNN, LSTM, Bi-LSTM, Convolutional-LSTM (C-LSTM), Hierarchical Attention Network (HAN), Conv-HAN, char-LSTM. For the CNN, LSTM, Bi-LSTM models, 100 dimensional pre-trained GloVe embedings were used and  HAN and Conv-HAN architectures employed 100 dimensional bidirectional GRU encoding. Their interpretation of the result is that no deep learning model is uniquely superior to others as no model performs consistently well on all three of their datasets. In addition, they note that while neural network-based models exhibit good accuracy and \( F_1 \)-score on a fairly large dataset (Combined Corpus), they exhibit overfitting on a small dataset (LIAR). Keya et al. in their paper - AugFakeBERT \cite{bib12} experimented with all three - traditional, deep learning and transformer based approach. For deep learning models they used CNN, LSTM, Bi-LSTM and CNN-LSTM with 128 dimensional trainable embeddings.


\subsection{Hybrid and Ensemble Approaches}\label{sec2.3}
Some research tried to merge multiple methods together to achieve better results. In this section, we discuss them. 


Ruchansky et al.~\cite{bib69} propose a hybrid model named \textbf{CSI}, consisting of three modules—\textit{Capture}, \textit{Score}, and \textit{Integrate}—and utilize two datasets: \textbf{TWITTER} and \textbf{WEIBO}. The first module employs an RNN to capture the temporal structure of user activity on a particular article, based on both user responses and textual content. User features ($x_u$) are embedded using Singular Value Decomposition (SVD), with ranks of 20 for Twitter and 10 for Weibo. The second module focuses on source characteristics derived from user activity, and the outputs of the first two modules are combined in the third module to determine the authenticity of an article. For Weibo, textual features ($x_\tau$) are extracted using \textit{doc2vec} applied to segmented text. The initial feature vector $x_t$ has a dimensionality of 122 for Twitter and 112 for Weibo. These raw features are projected into a lower-dimensional embedding space of 100 dimensions ($\tilde{x}_t$) before being passed into the RNN. Their hybrid CSI model was able to achieve 0.95 in accuracy and \( F_1 \)-score. 

The research team of Nasir et al. \cite{bib70} proposed a novel hybrid deep learning model for classifying fake news that integrates recurrent and convolutional neural networks. They used two publicly available datasets \textbf{FA-KES} and \textbf{ISOT} and employed 100 dimensional pre-trained GloVe embeddings. Both datasets have almost equal amounts of fake and real news. They were able to achieve an accuracy score of 0.60 in dataset-1 and 0.99 in dataset-2. 

The research of Ahmad et al. \cite{bib66} proposed ensembling multiple traditional ML models. They picked three publicly available English datasets which contain news from multiple domains. They explored various ensemble techniques such as bagging, boosting and voting classifier. Their first voting classifier is ensemble of LR,RF and KNN and second one consists of LR, linear SVM, and Classification and Regression Trees (CART). The bagging ensemble consists 100 DT and two boosting algorithms are XGBoost and AB. Comparing ensemble and individual learners, the ensemble learners have consistently outperformed the individual learners in their research.
	
\subsection{Transformer based Approaches}\label{sec2.4}
The field of natural language processing has been transformed by pre-trained transformer models like BERT (Bidirectional Encoder Representations from Transformers), which have displayed astounding performance on a variety of natural language processing tasks. Intricate associations and meanings are captured by these models as they learn contextualized word representations from huge datasets. Researchers have fine-tuned pre-trained transformer models successfully for the task of fake news classification.

Till now the only sufficiently large publicly available Bengali news dataset is BanFakeNews \cite{bib10} but the dataset is heavily skewed towards real news. Keya et al. \cite{bib12} suggested using text augmentation techniques to somewhat reduce the imbalance. Their research experiments with traditional, deep learning and transformer - all three approaches and they propose pre-trained transformer based approach. They fine tuned BERT base uncased architecture with balanced dataset and their build model - AugFakeBERT outperforms other methods by achieving an accuracy score of 92.45\%.

Saha et al. \cite{bib67} used two Bengali dataset and experimented with Pre-trained BERT transformer, LSTM with regularization, CNN, SVM and NB. They used two Bengali datasets separately for training and testing. In their research, BERT achieves 94\% \( F_1 \)-score which is superior to other methods explored.

A significant limitation in previous work lies in the common reliance on TF-IDF for feature extraction, which fails to capture contextual relationships within the text. Moreover, the limited availability of large and balanced fake news datasets often leads to the development of models susceptible to biases due to imbalanced training data. To overcome these critical shortcomings, our research aims to create a more comprehensive and balanced dataset through collection, translation, and augmentation, and to leverage the contextual understanding capabilities of the BERT model for improved fake news detection.

\begin{table}[h]
\centering
\caption{Notable studies on fake news detection using traditional methods, neural networks, hybrid, ensemble Approaches and transformers.}
\label{tab:my-table}
\resizebox{\columnwidth}{!}
{\begin{tabular}{|c|c|c|c|}
\hline
Type &
  Reference &
  Model used &
  Performance \\ \hline
\multirow{8}{*}{Traditional Approach} &
  \cite{bib61} &
  Traditional (SVM, LR, DT, MNB, KNN) &
  95\% accuracy (NB) \\ \cline{2-4} 
 &
  \cite{bib62} &
  SVM &
  \begin{tabular}[c]{@{}c@{}}100\% accuracy \\ using 5 features\end{tabular} \\ \cline{2-4} 
 &
  \cite{bib10} &
  SVM,LR, RF &
  91\% \( F_1 \) score(SVM) \\ \cline{2-4} 
 &
  \cite{bib63} &
  MNB, SVM &
  96.64\% accuracy(MNV) \\ \cline{2-4} 
 &
  \cite{bib64} &
  RF, PAC, MNB, SVM, LR, DT &
  \begin{tabular}[c]{@{}c@{}}93.8\% accuracy (PAC)\\ 93.5\% accuracy(SVM)\end{tabular} \\ \cline{2-4} 
 &
  \cite{bib68} &
  SVM,LR,MLP,RF,VEC,GNB &
  85.2\% accuracy(GNB) \\ \cline{2-4} 
 &
  \cite{bib72} &
  LR, Linear SVC, DT, RF, MNB, PAC &
  \begin{tabular}[c]{@{}c@{}}93.23\% accuracy\\  93\% \( F_1 \)-score\end{tabular} \\ \cline{2-4} 
 &
  \cite{bib73} &
  \begin{tabular}[c]{@{}c@{}}SVM, LR, MLP, RF, VEC, \\ GNB, MNB, AB , GB\end{tabular} &
  87\% accuracy \\ \hline
\multirow{4}{*}{Neural Network} &
  \cite{bib11} &
  CNN &
  96\% accuracy \\ \cline{2-4} 
 &
  \cite{bib12} &
  \begin{tabular}[c]{@{}c@{}}Traditional (LR, MNB, \\ DT, RF, SGD, SVM, KNN)\\ \\ DL (CNN, LSTM, Bi-LSTM, \\ CNN-LSTM, M-BERT)\\ \\ Transformer(BERT, M-BERT\end{tabular} &
  91\% accuracy(LSTM) \\ \cline{2-4} 
 &
  \cite{bib61} &
  \begin{tabular}[c]{@{}c@{}}DL (CNN, LSTM, Bi-LSTM, \\ C-LSTM, HAN, Conv-HAN,\\  char level C-LSTM)\end{tabular} &
  \begin{tabular}[c]{@{}c@{}}Dataset-1 - Conv HAN 59\% acc\\ Dataset-2 - C-LSTM 95\% acc\\ Dataset-3 - Bi-LSTM and C-LSTM \\ both 95\%\end{tabular} \\ \cline{2-4} 
 &
  \cite{bib75} &
  \begin{tabular}[c]{@{}c@{}}LSTM, CNN, \\ vanilla RNN, unidirectional LSTM\end{tabular} &
  \begin{tabular}[c]{@{}c@{}}91.08\% accuracy\\ (Bi-LSTM)\end{tabular} \\ \hline
\multirow{3}{*}{Hybrid and Ensemble} &
  \cite{bib70} &
  CNN + RNN &
  99\% accuracy \\ \cline{2-4} 
 &
  \cite{bib66} &
  \begin{tabular}[c]{@{}c@{}}Traditional ( \{LR,RF,KNN\},\\  \{LR,SVM,CART\}, \\ DT, AB, XGBoost)\end{tabular} &
  \begin{tabular}[c]{@{}c@{}}96.3\% accuracy \\ (XGBoost)\end{tabular} \\ \cline{2-4} 
 &
  \cite{bib69}\c &
  hybrid RNN &
  95\% accuracy \\ \hline
\multirow{2}{*}{Transformer} &
  \cite{bib12} &
  \begin{tabular}[c]{@{}c@{}}Traditional (LR, MNB, DT, \\ RF, SGD, SVM, KNN)\\ \\ DL (CNN, LSTM, Bi-LSTM, \\ CNN-LSTM, M-BERT)\\ \\ Transformer(BERT, M-BERT)\end{tabular} &
  92.45\% accuracy \\ \cline{2-4} 
 &
  \cite{bib67} &
  Transformer(BERT) &
  94\% accuracy \\ \hline
\end{tabular}}
\end{table}
\section{Corpus Description}\label{sec3}
In this section, we provide a brief description of the corpora that we used for training and test purposes. Throughout our research, we used a total of three corpora: two publicly available corpora and a manually collected corpus which we refer to as - `CustomFake'.

\begin{figure}[h]
\centering
\includegraphics[width=0.95\textwidth]{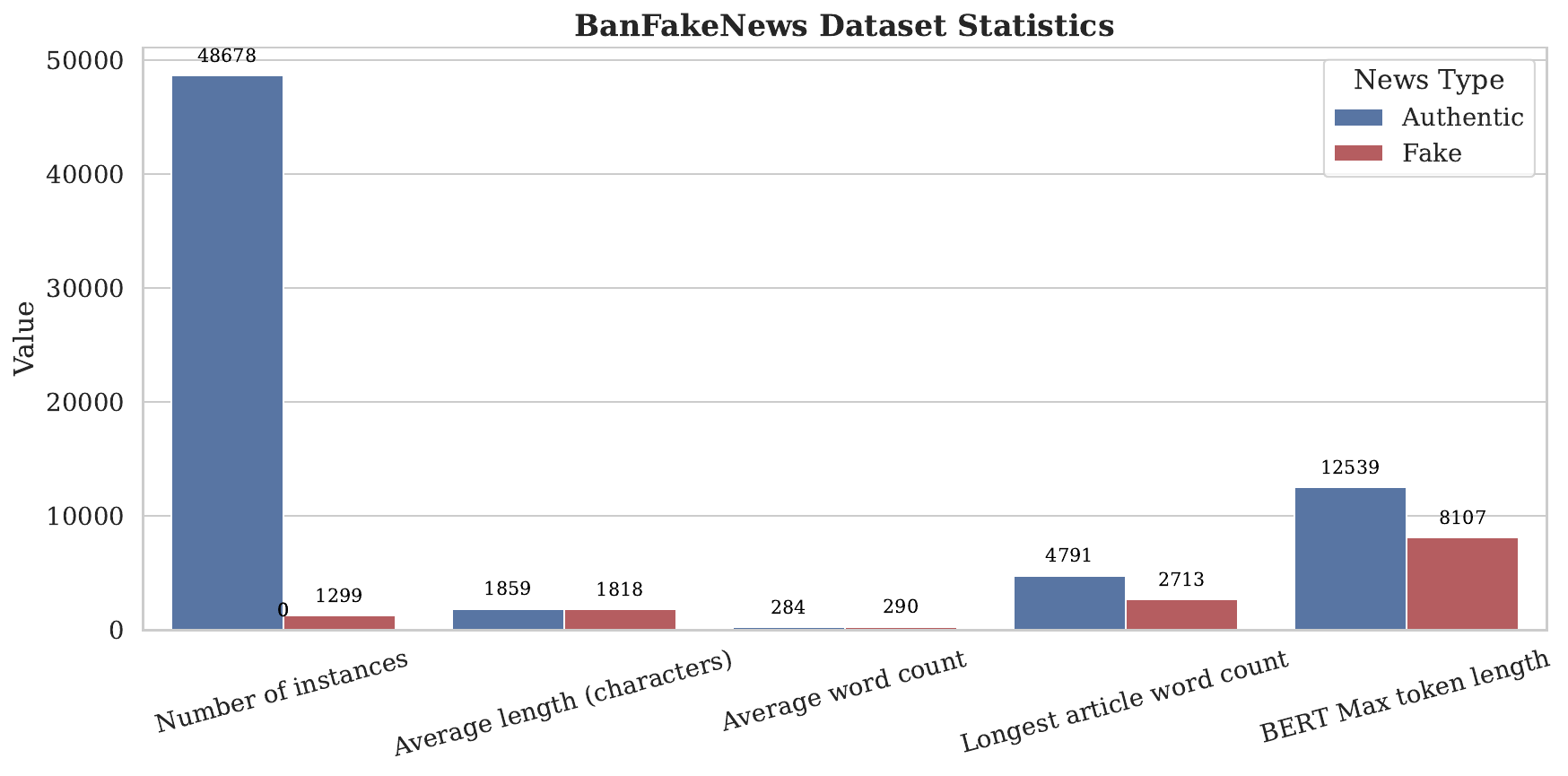}
\caption{Pipeline of the summarization process.}\label{fig:banfakenews_dataset_statistics}
\end{figure}

\noindent\textbf{(a) \underline{BanFakeNews Corpus}:} BanFakeNews \cite{bib10} dataset contains approximately 50,000 Bengali news articles among which about 48 thousand are of \textit{authentic} category and 1299 are of \textit{fake} category. The dataset contains seven attributes - \textit{articleID, domain, date, category, headline, content} and \textit{label}. The label here is binary where 0 denotes \textit{fake} class and 1 denotes \textit{authentic} class. For our research, we merged the headlines and their corresponding news content. We show a few statistics of the corpus in Figure \ref{fig:banfakenews_dataset_statistics}.

\noindent\textbf{(b) \underline{Fake News Detection Corpus}}: The second corpus that we utilized is an English fake news article corpus named Fake News Detection\footnote{\url{https://www.kaggle.com/datasets/jainpooja/fake-news-detection}}. This corpus has around 17000 articles related to \textit{news, politics, government, middle-east}. \textit{News} category articles are roughly 39\%. We only took the \textit{fake} articles of \textit{News} category and translated them to Bengali using the English-Bengali machine translation model provided by Hasan et al. \cite{bib54}. We refer to the translated fake news dataset as `TransFND' dataset. We removed empty cells and finally, we had 4309 \textit{fake} news articles which were necessary to curb the data imbalance of BanFakeNews corpus.

\noindent\textbf{(c) \underline{CustomFake Corpus}:} As discussed before, the BanFakeNews corpus lacks a sufficient number of fake news samples. To somewhat solve this problem, we collected fake news articles manually. We visited the fake news sources provided by Hossain et al. \cite{bib10} along with some new sources and collected a total of 102 new Bengali fake news articles and constructed a corpus we refer to as `CustomFake' corpus. We only use this corpus to evaluate the generalization performance of the proposed approaches. The fake news collection sources for the corpus are provided in Table \ref{tab:sources}.

\begin{table}[h]
\centering
\caption{Sources of the fake news articles collected for `CustomFake' dataset.}\label{tab:customfake corpus sources}
\label{tab:sources}
\resizebox{\columnwidth}{!}
{\begin{tabular}{@{}llll@{}}
\toprule
\textbf{Sources} & \textbf{Num of articles} & \textbf{Sources} & \textbf{Num of articles} \\ \midrule
Earki & 22 & bengali.news19 & 2 \\
jachai.org & 12 & Banladesh football ultra & 1 \\
hindustantimes.com  &  6  &  Qatar Airways  &  1 \\
bddailynews69.bd  &  6  &  The Daily star  &  1\\
nationalistview.com  &  5  &  Bangla Tribune  &  1\\
daily-star  &  4  &  songrami71  &  1\\
The Daily Star Bangla  &  3  &  sports protidin  &  1\\
zeenews  &  3  &  jamuna.tv  &  1\\
newschecker.in  &  3  & 71News24  &  1\\
DailyNews96.com  &  3  &  ntv  &  1\\
bangla.hindustantimes.com  &  3  &  Jugantor  &  1\\
Cinegolpo  &  2  &  roarmedia  &  1\\
banglainsider.com  &  2  &  Anandabazar  &  1\\
kalerkontho  &  2  &  awamiweb  &  1\\
Bangladesh24Online  &  2  & Aviation News  &  1\\
jagonews  &  2  &  Kolar kontho  &  1\\
priyobangla24.com  &  2  &  bengali.news18  &  1\\
bengali.news18.com  &  2 & & \\
\botrule 
\end{tabular}}
\end{table}

\subsection{Training Corpus}\label{train-corp-create}
For the training purpose, we have utilized both the \textit{BanFakeNews} and \textit{TransFND} corpora. Table \ref{tab: train instances} shows the number of instances in both of the training datasets. We describe the two training datasets below.
\begin{enumerate}[label=\textbf{({\alph*})}, wide, labelwidth=!, labelindent=0pt]
\item \textbf{\underline{Dataset 1}:} 
We acquired 4309 translated fake news from \textit{TransFND} corpus and we had 1299 fake news articles available in \textit{BanFakeNews} corpus. We merged these articles which resulted in a total number of 5608 Bengali \textit{fake} news articles. We randomly selected an equal number of \textit{authentic} news from BanFakeNews. We separated 1200 news articles for testing which has an equal number of fake and real news. Thus `Dataset 1' was left with 10,016 news articles, where 5008 is fake and 5008 is authentic. 


\item \textbf{\underline{Dataset 2}:}
Augmentation techniques are used for reducing the imbalance of the minority class brought on by insufficient samples. To create the second training dataset, in this work, we applied two augmentation techniques - \textit{token replacement} and \textit{paraphrasing}. We augmented 1299 \textit{fake} news articles of \textit{BanFakeNews} two times and we got a total of 3897 ($1299\times2$ and original 1299) \textit{fake} news articles. We performed experimentation with 3507 \textit{fake} instances and an equal number of real news. We also experimented by taking four augmentations (using \textit{back translation} along with other two to augment) for a single fake instance and observed that most augmentations only showed subtle changes which resulted in over fitting problem during training. Therefore, we created `Dataset 2' by augmenting a single fake instance using \textit{token replacement} by leveraging BanglaBERT\footnote{\url{https://huggingface.co/sagorsarker/bangla-bert-base}} and \textit{paraphrasing} by leveraging BanglaT5\footnote{\url{https://huggingface.co/csebuetnlp/banglat5_banglaparaphrase}}. 
\end{enumerate}
\begin{table}[htbp]
    \begin{minipage}[t]{0.45\linewidth}
        \centering
        \begin{tabular}{|c|c|c|c|}
            \hline
            \textbf{Corpus} & \textbf{Auth} & \textbf{Fake}\\ \hline
            Dataset 1     & 5008   & 5008 \\
            Dataset 2            & 3507   & 3507 \\ \hline
        \end{tabular}
        \subcaption{Training} \label{tab: train instances}
    \end{minipage}
    \hspace{0.5cm}
    \begin{minipage}[t]{0.45\linewidth}
        \centering
        \begin{tabular}{|c|c|c|}
            \hline
            \textbf{Corpus}     & \textbf{Auth} & \textbf{Fake} \\ \hline
            Test Dataset-1   & 600   & 600  \\
            Test Dataset-2     & 2000  & 2000  \\
            Test Dataset-3        & 102   & 102  \\ \hline
        \end{tabular}
        \subcaption{Test} \label{test-instances}
    \end{minipage}
\caption{Number of instances in training and test datasets.}
\label{train-test instances}    
\end{table}

\subsection{Test Corpus}\label{test-corp-create}
To evaluate the performance of our developed models, we created three different test corpora. Table \ref{test-instances} shows the data distributions in each test corpus. We describe the three test datasets below.
\begin{enumerate}[label=\textbf{({\alph*})}, wide, labelwidth=!, labelindent=0pt]
    \item \textbf{\underline{Test Dataset-1}:} In this test dataset, there are a total of 1200 instances where 600 instances are \textit{fake} and the rest 600 are \textit{authentic} news articles. The instances of test dataset are separated from the training `Dataset 1' where the instances were from both the `BanFakeNews' and `TransFND' corpora.     
    \item \textbf{\underline{Test Dataset-2}:} To create the second test dataset, we randomly pick 2000 \textit{fake} articles from 4309 `TransFND' dataset. Then we randomly pick 2000 \textit{authentic} articles from `BanFakeNews' dataset. We concatenate \textit{fake} and \textit{authentic} data instances and shuffle them. While creating this test dataset, we made sure that the 2000 authentic instances were not used in the training dataset. Moreover, we only use this test dataset to evaluate the performance of our developed models while training with `Dataset 2' because there are no data instances of `TransFND' in the second training dataset. 
    \item \textbf{\underline{Test Dataset-3}:} We use the `CustomFake' corpus as the third test dataset where we manually collected 102 new \textit{fake} news articles from various news websites. We pair them with 102 randomly selected \textit{authentic} articles from BanFakeNews corpus. While creating this test dataset, we made sure that the 102 \textit{authentic} news articles were not used in the training datasets.
\end{enumerate}

\subsection{Rationale for Selecting Dataset Combinations}
The rationale for selecting such combinations for training and testing datasets is of three folds - \textit{ensuring balanced training, evaluating robustness as well as generalization capability,} and \textit{leveraging diverse data characteristics}. By creating balanced training datasets (Dataset 1 and Dataset 2) with an equal number of real and fake news samples, we aim to prevent the classifier from being biased towards the majority class (real news) during training. In both training datasets and all test datasets, the authentic news articles are from `BanFakeNews' dataset \cite{bib10}. Selecting authentic news articles exclusively from the `BanFakeNews' dataset ensures consistency in the source of authentic data between training and testing. This consistency helps to isolate the impact of the model's ability to distinguish between fake and real news from any potential variations in the quality or characteristics of the authentic news sources. Mixing fake news articles from different datasets \textit{(TransFND, BanFakeNews)} with real news articles from a single dataset \textit{(Banfakenews)} for training helps the model to generalize better on unseen data. In addition, by paraphrasing and replacing tokens in the fake news articles, we artificially create more training data with linguistic variations. This helps the model become robust to slight wording changes often seen in fake news. Different datasets might contain fake news articles with varying characteristics, sources, and topics, thereby making the model more robust and generalizable. The reason for selecting fake articles from the \textit{TransFND} dataset and authentic articles from the \textit{BanFakeNews} dataset for Test Dataset-2 is to evaluate the model's cross-dataset generalization ability. This is important because fake news can vary significantly in terms of style, content, source, and a robust detection model should be able to identify fake news regardless of its origin. We create Test Dataset-3 with the aim of assessing how well the fine-tuned models perform on real-life fake news articles in terms of their ability to handle variations and generalizations. As we chose fake and genuine articles randomly from the datasets, we ensured that there was no overlap between articles in the test and training datasets.

\section{Methodology}\label{sec4}
In this section, we present a detailed overview of our proposed methodology. 

\subsection{Proposed Approaches}\label{prop-approach}
\begin{figure}
     \centering
     \begin{subfigure}[b]{0.4\textwidth}
         \centering
         \includegraphics[width=\textwidth]{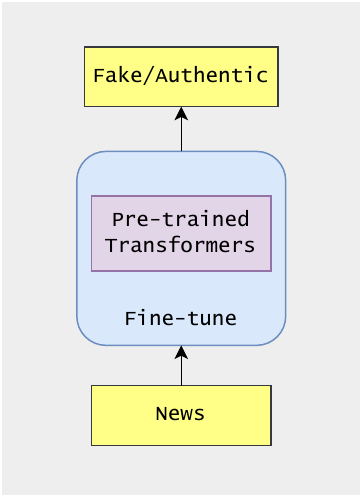}
         \caption{Approach 1}
         \label{fig:baseline}
     \end{subfigure}
     \begin{subfigure}[b]{0.4\textwidth}
         \centering
         \includegraphics[width=\textwidth]{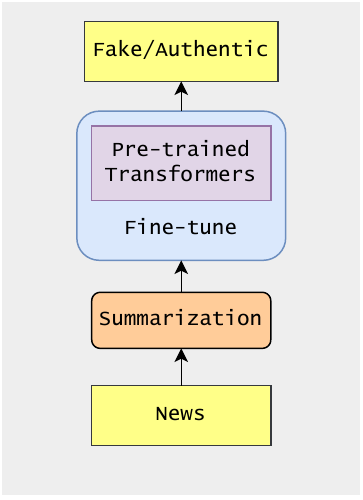}
         \caption{Approach 2}
         \label{fig:baseline summary}
     \end{subfigure}\\
     \begin{subfigure}[b]{0.4\textwidth}
         \centering
         \includegraphics[width=\textwidth]{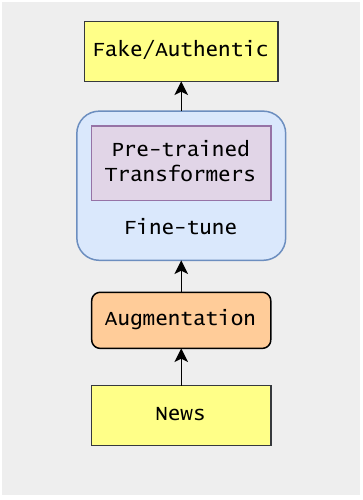}
         \caption{Approach 3}
         \label{fig:baseline augmented}
     \end{subfigure}
     \begin{subfigure}[b]{0.4\textwidth}
         \centering
         \includegraphics[width=\textwidth]{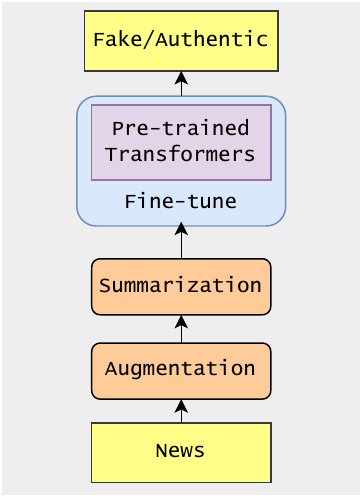}
         \caption{Approach 4}
         \label{fig:augmented summary}
     \end{subfigure}
        \caption{Schematic diagram of the proposed approaches.}
        \label{fig:proposed_approaches}
\end{figure}

We have explored four distinct approaches. The first approach involves fine-tuning pre-trained transformers \cite{bib52} that serve as the baseline models. In the second approach, we employ summarized news articles to fine-tune these baseline models. The third approach involves fine-tuning the baseline models using both actual and augmented news articles. The fourth approach involves utilizing both summarized actual and augmented news articles for fine-tuning the baseline models. Figure \ref{fig:proposed_approaches} illustrates all four approaches. We provide detailed descriptions of each approach below.

\noindent\textbf{(a) \underline{Approach 1: Fine-tuning Transformers.}}\\
In this approach, we have performed fine-tuning on five pre-trained language models using `Dataset 1'. We followed the standard fine-tuning procedure for sequence classification task. Each text from the dataset is tokenized, [CLS] token is added to represent the beginning of a sequence. Each text is padded or truncated to the maximum length of 512. During fine-tuning, through back propagation, the weights of the classification layer and some of the upper layers of the language model are updated. In this approach, we did not introduce summarization or augmentation. Figure \ref{fig:baseline} represents the fine-tuning approach.

\noindent\textbf{(b) \underline{Approach 2: Fine-tuning Transformers via Summarization.}}\\
Figure \ref{fig:baseline summary} represents this approach. In this approach, we began by summarizing the news articles from `Dataset 1'. Then, we fine-tuned the pre-trained language models using these summarized news articles. The reason for implementing summarization is that pre-trained transformers can only handle up to 512 tokens, and news articles are usually longer than that limit. As a result, it is possible that the classification model could potentially miss crucial words in a news article, which are vital for distinguishing fake news from authentic ones. Therefore, we felt the need to condense the knowledge within the entire news articles as much as possible by summarizing, allowing the classifier model to focus on the most important information. We discuss the details of the summarization procedure in the following section.

\noindent\textbf{(c) \underline{Approach 3: Fine-tuning Transformers via Augmentation.}}\\
Figure \ref{fig:baseline augmented} provides a visual representation of this approach. Initially, we constructed `Dataset 2' where we only augmented the fake news articles of the \textit{BanFakeNews} dataset leaving the authentic news articles as they are. We then fine-tuned the pre-trained language models using Dataset 2. The purpose behind employing augmentation techniques was to introduce variations and noise into the fake news data. This was achieved by incorporating random perturbations such as token replacements. Text augmentation methods such as paraphrasing was utilized to preserve the semantics of the original text while presenting it in a different manner. This aspect holds particular significance in fake news detection as subtle alterations in how information is presented can significantly impact the ability of the model to identify fake news. Pre-trained language models when fine-tuned on augmented data, can learn more diverse representations which enhances their capability to generalize across different contexts and sources of fake news.

\noindent\textbf{(d) \underline{Approach 4: Fine-tuning via Summarization and Augmentation.}}\\ 
The primary concept behind this approach is to examine how summarizing the augmented data influences the process of detecting fake news as illustrated in Figure \ref{fig:augmented summary}. Initially, we condensed the content of 'Dataset 2' through summarization and then we performed the fine-tuning process using this summarized dataset. By distilling the augmented information into concise summaries, the model becomes capable of learning from these condensed representations, enabling it to identify shared characteristics and distinctive factors of fake news. This can enhance the capability of the model to detect fake news from previously unseen sources or variations of fake news.

\subsection{Language Models}
In this study, we fine-tuned few Bidirectional Encoder Representations from Transformers (BERT) \cite{bib51} variant language models pre-trained on Bengali texts in order to detect fake news. BERT works bidirectionally and employs a Transformer \cite{bib52} encoder architecture that includes multiple layers of self-attention mechanisms and feed-forward neural networks. By utilizing these layers, BERT can effectively understand the contextual connections between words within a sentence and produce high quality representations (embeddings) for each word. These embeddings enable BERT to make precise predictions by combining them with a task-specific architecture and further training. We have performed fine-tuning on five different variants of BERT models among which three are multi-lingual BERT models. We show the architectures of all the fine-tuned models in Table \ref{tab:architecture}.\\

\noindent\textbf{(a) Multilingual BERT based.} Multilingual BERT (mBERT) language model was pre-trained on 104 languages using masked language modeling (MLM) objective. We fine-tuned the pre-trained base mBERT checkpoint\footnote{\url{https://huggingface.co/bert-base-multilingual-cased}} available in Hugging Face Transformers Library \cite{wolf2020transformers}. We also fine-tuned two other available mBERT checkpoints pre-trained on Bengali fake news detection task. We refer to those models as TM-mBERT\footnote{\url{https://huggingface.co/Tahsin-Mayeesha/bangla-fake-news-mbert}} and DB-mBERT\footnote{\url{https://huggingface.co/DeadBeast/mbert-base-cased-finetuned-bengali-fakenews}} in this study. TM-mBERT and DB-mBERT were developed by fine tuning the original base mBERT model using the \textit{BanFakeNews} dataset.\\

\noindent\textbf{(b) BERT based.} To detect fake news, we fine-tuned two BERT variant Bengali language models: BanglaBERT and BanglaBERT Base. BanglaBERT \cite{bib58} is pre-trained on a vast collection of Bengali texts using ELECTRA pre-training objective. On the other hand, BanglaBERT Base is pre-trained over two corpora: Bengali common crawl corpus and Bengali Wikipedia Dump Dataset. We fine-tuned the checkpoints\footnote{\url{https://huggingface.co/csebuetnlp/banglabert}}\hochkomma\footnote{\url{https://huggingface.co/sagorsarker/bangla-bert-base}} available in Hugging Face Library.

\begin{table}[h]
\centering
\caption{Architectures of the pre-trained transformer models.}
\label{tab:architecture}
\resizebox{\textwidth}{!}
{\begin{tabular}{lccccc} 
\hline
\multicolumn{1}{c}{\textbf{}} & \begin{tabular}[c]{@{}c@{}}\textbf{TM}\\\textbf{~mBERT~}\end{tabular} & \begin{tabular}[c]{@{}c@{}}\textbf{Bangla}\\\textbf{BERT}\end{tabular} & \begin{tabular}[c]{@{}c@{}}\textbf{~mBERT~}\\\textbf{Base}\end{tabular} & \begin{tabular}[c]{@{}c@{}}\textbf{DB}\\\textbf{~mBERT~}\end{tabular} & \begin{tabular}[c]{@{}c@{}}\textbf{BanglaBERT}\\\textbf{~Base}\end{tabular}  \\ 
\hline
Embedding size                & 768                                                                   & 768                                                                    & 768                                                                     & 768                                                                   & 768                                                                          \\
No. of attention heads        & 12                                                                    & 12                                                                     & 12                                                                      & 12                                                                    & 12                                                                           \\
No. of hidden layers          & 12                                                                    & 12                                                                     & 12                                                                      & 12                                                                    & 12                                                                           \\
Max positional embedding      & 512                                                                   & 512                                                                    & 512                                                                     & 512                                                                   & 512                                                                          \\
Vocabulary size               & 119547                                                                & 32000                                                                  & 119547                                                                  & 119547                                                                & 102025                                                                       \\
\hline
\end{tabular}}
\end{table}

\subsection{Augmentation}\label{aug}
\begin{figure}[h]%
\centering
\includegraphics[scale=0.5]{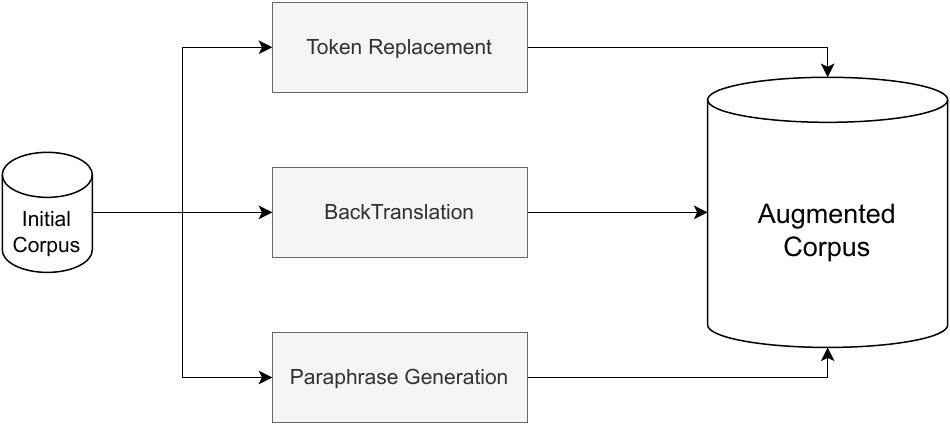}
\caption{Pipeline of the augmentation process.}\label{fig-augpipe}
\end{figure}

A popular technique in natural language processing (NLP) to increase the quantity of training data samples is to use \textit{text augmentation}. Among several available techniques, we chose three different types of augmentation techniques - \textit{token replacement, back translation} and \textit{paraphrase generation}. Figure \ref{fig-augpipe} illustrates the augmentation pipeline used in this study. We show the application of these augmentation techniques by taking an example sentence in Table \ref{tab:aug example table}.\\

\begin{enumerate}[label=\textbf{({\alph*})}, wide, labelwidth=!, labelindent=0pt]
\item \textbf{\underline{Back translation}:} Back translation method works as follows: we take some sentences (e.g. in Bengali) and translate to another language (e.g English). Then we translate the English sentences back to Bengali sentences. The final Bengali sentences will have subtle changes than the original Bengali sentences. We used the publicly available \verb|bnaug|\footnote{\url{https://github.com/sagorbrur/bnaug}} library to perform back translation. This technique preserves the semantic meaning of the original text while introducing syntactic and lexical variations. It is particularly effective for low-resource languages like Bengali, where other augmentation methods may struggle to maintain linguistic integrity \cite{pingle2023robust}.

\item \textbf{\underline{Paraphrase generation}:} Paraphrase generation preserves the meaning of the sentence while changing the grammar and word choices. There are two ways to achieve this - Rule Based and machine learning based. In rule based method rules are created manually for example randomly selecting one or more words in sentences and changing them with words. This may also include changing active voice into passive, changing parts of speech etc. On the other hand, in machine learning based approach, paraphrase is automatically generated from data. Pre-trained seq2seq (sequence-to-sequence) language models are capable of generating paraphrases automatically. The \verb|bnaug| library we used uses a BanglaT5 checkpoint\footnote{\url{https://huggingface.co/csebuetnlp/banglat5_banglaparaphrase}} provided by Akil et al. \cite{banglaparaphrase} specifically pre-trained on Bengali paraphrasing task.\\

\begin{table}
\centering
\caption{Different augmentation techniques applied to a sample text.}
\label{tab:aug example table}
\begin{tabular}{ccc} 
\hline
\textbf{Text Process} & \textbf{Bengali Text}          & \textbf{English Translated Text}                                                           \\ 
\toprule
Original              & {\bng murigr Hamlay eshyal inHoto}     & Fox killed by chicken attack                                                               \\
Token replaced        & {\bng murigr AivJaen eshyal inHoto}     & Fox killed in chicken raid                                                                 \\
Back translated       & {\bng murigr AakRomoeN eshyal inHoto}     & Fox killed by chicken onslaught                                                            \\
Paraphrased           & {\bng murigr AakRomoeN ishyal inHoto Hoy} & \begin{tabular}[c]{@{}c@{}}The fox was killed by the \\attack of the chicken\end{tabular}  \\
\hline
\end{tabular}
\end{table}

\item \textbf{\underline{Token replacement}:} The last technique we used was token replacement using random masking. BERT models are pre-trained on a huge volume of texts using `Masked Language Modelling' (MLM) objective where the model has to predict masked words based on a context. This can be used for augmentation where some tokens are randomly masked in a sentence and a language model has to predict the token for that mask. For token replacement, we used \verb|nlpaug| library\footnote{\url{https://github.com/makcedward/nlpaug}} with two BERT variations: BanglaBERT \cite{bib58} and ShahajBERT\footnote{\url{https://huggingface.co/neuropark/sahajBERT}}.
\end{enumerate}

\subsection{Summarization}\label{summarization}
\begin{figure}[h]
\centering
\includegraphics[width=0.95\textwidth]{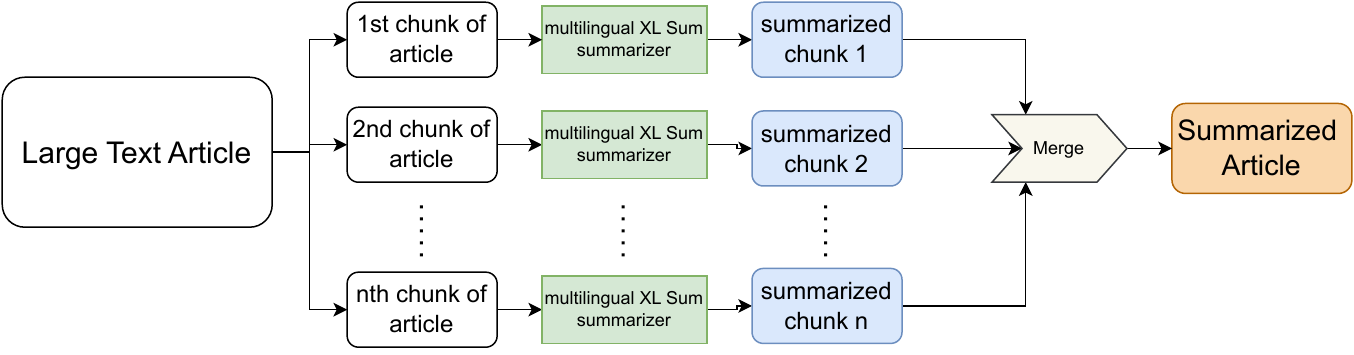}
\caption{Pipeline of the summarization process.}\label{fig-sumpipe}
\end{figure}

To tackle the issue of BERT models not being able to take token sequence length greater than 512, we summarized our news articles which exceeded the limit. Our hypothesis was that rather than taking top truncated part of the news, if summarized news was used for training, the models should perform better. As we know, sometimes large news articles start with irrelevant information. To summarize articles in our corpus, we used a mT5 checkpoint\footnote{\url{https://huggingface.co/csebuetnlp/mT5_multilingual_XLSum}} pre-trained on multilingual summarization task provided by Hasan et al. \cite{bib55} which generates the abstractive summarization. A base mT5 model was fine-tuned using one million annotated article-summary pairs covering 44 languages including Bengali. To make sure that summary contained texts from all parts of the article, we divided large articles into chunks, summarized those chunks and then merged the chunks to create a summarized text. The summarization pipeline is depicted in Figure \ref{fig-sumpipe}. This was necessary as some of the articles were very large, even as large as 19,000 tokens. An example of summarization is provided in Table \ref{tab5}.

\begin{table}
\centering
\caption{Example of summarization of a sample news article instance. \textit{B} represents Bengali and \textit{E} represents the corresponding English translation.}
\label{tab5}
\begin{tabular}{|p{0.45\textwidth}|p{0.45\textwidth}|}
\hline
\textbf{Article} & \textbf{Summary} \\ 
\hline
{\bng
ekakaekalar ekamol paniiyet imishRto Hoet Jaec/ch  gaNNja! ekaka-ekala Ebar panii-yet gajNNar inr/Jas eJag kret Jaec/ch. ekaka-ekala blech, gaNNja sNNGishLSh/T paniiy bajar pr/JebkKN krech tara. kanaDaibhit/tk ibEnEn bLumbar/g iTiv EI tthY jaineyech. EcharhaO s/kaIinUj, mYaeshbl, Eibis inUj sH ibeshWr pRthm sairr gNmadhYm E khbr pRkash kerech. bla Hec/ch, s/thaniiy Ut//padk 'Aerara kYanaibs' Er soeNG/g gaNNjar sWadJuk/t ekaml paniiy Ut//padenr ibShey Aelacna krech ekaka-ekala. teb gRaHkedr madkask/t koret ny, taedr shariiirk Jn/tRNa laghb kraI paniiy oitirkariiedr lkKY. ekaka-ekala Ek ibbrRitet jaineyech, Aenk Ut//padekr meta AamraO pr/J-ebkKN korich eJ, ekaml paniiy oitirr ekKetR nn-saIekaAYaik/Tbh kYanaibiDOl ba ict/to Uet/tijt ker na Emn gaNNja jatiiy dRebYr bYbHar kotTa jonipRyota paec/ch. kYanaibiDOl kYanaibs ba gaNNjar EkiT Upadan, Ja pRdaH, bYtha ba ikhcuinr icikt//sar ekKetR Aaramdayk Het paer EbNNG Er ekaena ict/to Uet/tjk pRvab enI. gaNN-jar ibenadnmuulk bYbHar AaInot oibdh kret Jaec/ch kanaDa. icikt//sa kaej AboshYo Aenk Aaeg ethekI gaNNja oibdho kanaDay. ekaka-ekalar E isd/dhaen/tr foel kanaDay goerh UeThech ibshal Aakaerr gaNNja iSl/po AnYidek, Ut//padonkarii sNNGs/tha emalson kuros BRuiyNNG boelech, tara HaIeDRaepaethkYair soNNGeJajn ker gaNNja iniShk/to paniiy oitir koreb. IetamoedhY ibshWokhYato koerana ibyar oitirkarii soNNGs/tha kones/TelSan bRYan/Ds gaNN-ja Ut//padonkarii pRitSh/Than kYaenaip egRaethr Opor car ibilyn Dolar ibinyog koerech. ekakaekala Aar Aerara'r ANNGSiidairetWr foel gaNNjar paniieyr bajaer pRothom non-AYalekaHilk paniiy iHeseb JatRa shuru Hoeb ekak'Er. jana egech, Aerara'r soeNG/g ekaka-ekala'r Aaelacona Aenkdur AgRosor HoelO curhan/to ekaena cuik/t Hyin. suutRiT jaineyech, EI paniiyiT shudhu AbsadI duur koreb na , soetjota laevO soHoyota koreb. Aalada Ek ibbrRitet Aerara jaineyech cuik/t cuurhan/to na HOya por/Jon/to tara E ibShoey ibs/tairto ikchu janaeb na. toeb tara boelech, gaNNja iniShk/t paniieyr bajaer pRoebsh korar ibShoey Aerara JoethSh/To AagRoHii roeyech. }

& {\bng [B] ekakaekalar ekamol paniiyet imishRto Hoet Jaec/ch  gaNNja! Emon khobor pRokash koerech ibeshWr pRothom sairr  goNomadhYom  EboNNG samaijk eJagaeJag madhYoemr ebsh koeykjon ibEnEn inUj is/TRimNNG Er soHoeJagii soNNGgoThon 'ekaka-ekala'. Ekhobor paOya egech, Juk/toraeSh/TRr EkiT Ut//padoekr soeNG/g Aaelacna korar por tara bolech eJ, EI koeranavaIraesr soNNG-kRomoN eThkaet Juk/toraeSh/TRr gaNNja Ut//padonkarii soNNGs/ha  ekaka-ekala Aar kYanaDay notun EkiT non-AYalekaHilk paniiyo oitir korar ibShey Aerara'r ANNGshiidairtW paOyar isd/dhan/t ineyech edshiTr sWas/hYo mon/tRoNaloeyr suutRo. ikn/tu EI cuik/tet ekak'Er soeNG/g Aelacona AgRosor HOyar por tara bolech, Ekhon ethek.}

\hfill \break

[E] Cannabis is going to be mixed with Coca-Cola's soft drinks! Such news has been revealed by the world's leading mass media and several social media, affiliated organization of BNN News Streaming - 'Coca-Cola'. It has been reported that, after discussing with a producer in the United States, they say that "to prevent the spread of this coronavirus, the American cannabis production company Coca-Cola and Aurora have decided to partner up in Canada to develop a new non-alcoholic drink, according to the country's Ministry of Health. But after the progress of negotiations with Coke on this deal, they say, from now on.
\\
\hline
\end{tabular}
\end{table}

\section{Experiments}\label{sec5}
In this section, we conduct a comprehensive evaluation to report the performance and effectiveness of the proposed approaches.

\subsection{Evaluation Metrics}\label{eval-metrics}
For the purpose of evaluation, we have employed various standard metrics including accuracy, precision, recall, MCC (Matthews Correlation Coefficient), and ROC-AUC score. We provide a brief explanation of these metrics below. In the following equations, TP represents True Positives, TN represents True Negatives, FP represents False Positives, and FN represents False Negatives.\\

\noindent\textbf{Accuracy:} The ratio of correctly classified data and total number of data. 
\begin{equation}
\text { accuracy }=\frac{T P+T N}{T P+T N+F P+F N}
\end{equation}

\noindent\textbf{Precision:} The proportion of positive cases that were correctly classified.
\begin{equation}
\text { precision }=\frac{T P}{T P+F P} 
\end{equation}

\noindent\textbf{Recall:} The ratio of correctly classified positives and actual positives. 
\begin{equation}
\text { recall }=\frac{T P}{T P+F N}
\end{equation}

\noindent\textbf{\( F_1 \) Score:} The \( F_1 \) Score is the harmonic mean of precision and recall. In our binary classification task, we evaluate model performance by computing the macro average of metrics such as the \( F_1 \) score. The macro average is the arithmetic mean of the \( F_1 \) scores calculated independently for each class. 
\begin{equation}
 \( F_1 \) = 2 \cdot \frac{\text{precision} \cdot \text{recall}}{\text{precision} + \text{recall}}   
\end{equation}



\noindent\textbf{MCC:} The Matthews correlation coefficient (MCC) is least influenced by imbalanced data. It is a correlation coefficient between the observed and predicted classifications. The value ranges from -1 to +1 with a value of +1 representing a perfect prediction, 0 as no better than random prediction and -1 the worst possible prediction.
\begin{equation}
M C C=\frac{T P \times T N-F P \times F N}{\sqrt{(T P+F P)(T P+F N)(T N+F P)(T N+F N)}}
\end{equation}

\noindent\textbf{ROC-AUC Score:} The ROC or the Receiver Operating Characteristic curve calculates the sensitivity and specificity across a continuum of cutoffs. An appropriate balance can be determined between sensitivity and specificity using the curve. AUC is simply the area under the ROC curve. An area of 1 represents a perfect model and an area of 0.5 represents a worthless model.

\subsection{Experimental Setup}\label{sec:exp setup}
To carry out all the experiments, we used Google Colaboratory platform with 12.7GB system RAM, 15GB google Tesla T4 GPU RAM and 78.2GB disk space. To process and prepare data we used \verb|pandas| and \verb|numpy| library. For implementation purpose, we used \verb|python| language and transformer models were implemented using \verb|Pytorch| framework. During the training and validation stages, we maintained an $85:15$ ratio on the training datasets. For testing purposes, we created three separate test datasets.

\subsection{Hyper-parameters Tuning}
\begin{figure}[h]
\centering
\includegraphics[width=0.95\textwidth]{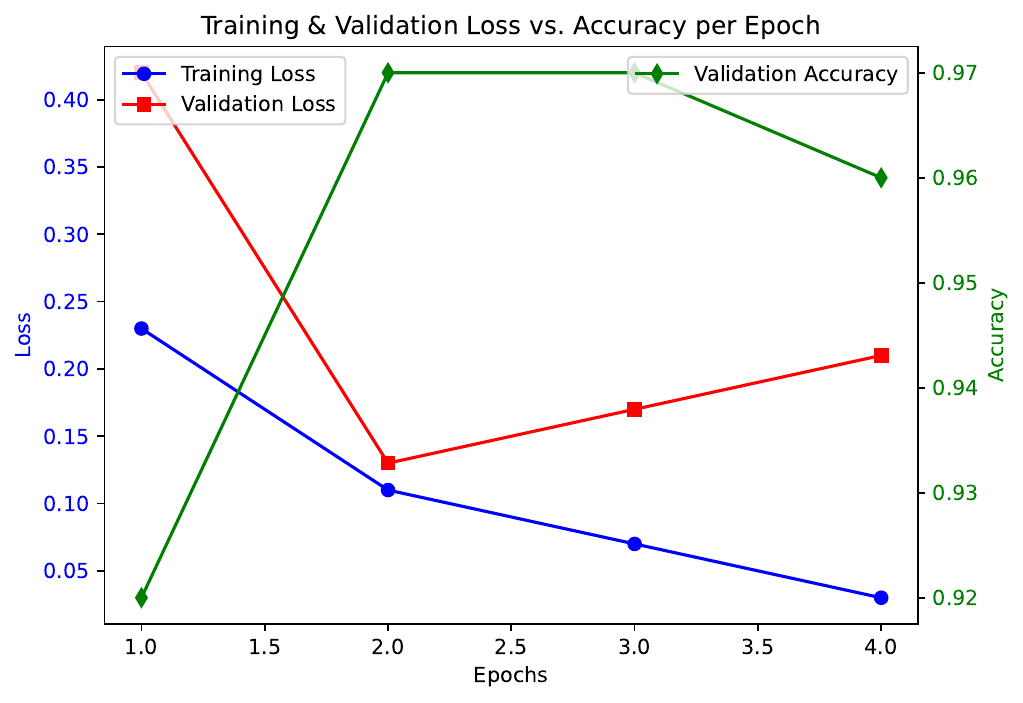}
\caption{Training and validation loss vs Accuracy for TM-mBERT.}\label{fig:training_validation_curves}
\end{figure}

Hyper-parameters are parameters that are set before the training process begins such as the learning rate, batch size, loss function, activation function, the number of epochs etc. The hyperparameters were fine-tuned to achieve the best performance on the validation set. Table \ref{tab:hyper} presents the final hyperparameter settings used across all experiments, while Figure \ref{fig:training_validation_curves} illustrates the relationship between training and validation loss versus accuracy. As our train and test datasets are balanced, we consider accuracy and \( F_1 \)-score as the primary metrics for assessing the superiority of the models. Since all classes have an equal number of samples, we specifically focus on the macro average. Moreover, we take note of other scores such as precision, recall, MCC, and ROC-AUC to gain a more comprehensive understanding of the performance of each model.

\begin{table}[h]
\centering
\caption{Hyper-parameter settings}
\label{tab:hyper}
\begin{tabular}{@{}ll@{}}
    \toprule
        \textbf{Hyper-parameter} & \textbf{Value} \\ 
    \midrule
        Maximum Sequence Length & 512 \\ 
        Epochs & 4 \\ 
        Batch Size & 16 \\ 
        Activation Function & Gelu, Softmax \\ 
        Learning Rate & 2e-05 \\ 
        Optimizer & AdamW \\
        Loss Function & Binary Cross Entropy\\
    \botrule
\end{tabular}
\end{table}


\subsection{Evaluation}\label{sec:Results}
The research questions outlined in Section \ref{sec1} are discussed in detail below.

\vspace{1em}

\noindent\fbox{%
    \parbox{\textwidth}{%
        \textbf{RQ1: How does fine-tuning without summarization and augmentation perform when translated news (English-Bengali) are combined with Bengali news articles for the task of fake news classification?}
    }%
}\\

In \textit{Approach 1}, we focused on fine-tuning all the transformer models without the use of summarization or augmentation. During training, we incorporated translated fake news (English-Bengali) alongside Bengali news articles. The performance of these fine-tuned models was evaluated using `Test Dataset-1' and `Test Dataset-3'. We show the performance of the fine-tuned models on `Test Dataset-1' and `Test Dataset-3' in Table \ref{restab: Test Dataset-1 test} and \ref{restab: Test Dataset-3 test} respectively. The fine-tuned models in \textit{Approach 1} exhibited impressive performance on `Test Dataset-1', with the BanglaBERT model achieving the lowest accuracy of 88\%. The mBERTBase model achieved the highest accuracy of 92\%, while the other three models also scored above 90\%. On the other hand, the BanglaBERT Base model attained the highest accuracy score of 80\% on `Test Dataset-3'. This custom test dataset, designed to assess generalization performance, was completely unfamiliar to the fine-tuned models. The results obtained from \textit{Approach 1} indicate that it is feasible to develop a robust Bengali fake news detection model without relying on summarization or augmentation techniques, particularly when a substantial amount of fake data is available. In our case, we achieved this by translating fake English news into Bengali, thereby increasing the number of fake news instances. Overall, the BanglaBERT Base model demonstrated better performance on both test datasets.

\begin{table}[h]
\centering
\caption{Performance comparison across all the approaches on the first test dataset `Test Dataset-1'. \textit{Inference} represents performing classification without any training or fine-tuning.}\label{restab: Test Dataset-1 test}
\begin{tabular}{@{}llllllll@{}}
\toprule
    \textbf{Method}&\textbf{Model} & \textbf{A}    & \textbf{P}    & \textbf{R}    & \(\mathbf{F}_1\)   & \textbf{MCC}   & \textbf{ROC}   \\
\midrule    
&TM-mBERT           &   0.11 & 0.09 & 0.11 & 0.10 & -0.80 & 0.61 \\ 
&BanglaBERT &           0.50 & 0.25 & 0.5  & 0.33 & 0.00  & 0.50 \\ 
Inference & mBERT-Base & 0.53 & 0.54 & 0.53 & 0.47 & 0.07  & 0.41 \\ 
& DB-mBERT            & 0.11 & 0.09 & 0.11 & 0.10 & -0.79 & 0.57 \\ 
& BanglaBERT-Base       & 0.52 & 0.66 & 0.52 & 0.39 & 0.12  & 0.50 \\ 
\midrule
  & TM-mBERT           & 0.91 & 0.91 & 0.91 & 0.91 & 0.81 & 0.59 \\ 
& BanglaBERT & 0.88 & 0.89 & 0.88 & 0.88 & 0.77 & 0.59 \\ 
Approach 1 & mBERT-Base              & 0.92 & 0.92 & 0.92 & 0.92 & 0.84 & 0.48 \\ 
& DB-mBERT          & 0.91 & 0.91 & 0.91 & 0.91 & 0.82 & 0.55 \\ 
& BanglaBERT-Base      & 0.90 & 0.90 & 0.90 & 0.90 & 0.80 & 0.49 \\ 
\midrule
&TM-mBERT           & 0.84 & 0.87 & 0.84 & 0.84 & 0.71 & 0.55 \\ 
&BanglaBERT & 0.76 & 0.80 & 0.76 & 0.75 & 0.56 & 0.56 \\ 
Approach 2 & mBERT-Base              & 0.87 & 0.87 & 0.87 & 0.87 & 0.74 & 0.51 \\ 
& DB-mBERT           & 0.88 & 0.89 & 0.88 & 0.88 & 0.77 & 0.57 \\ 
& BanglaBERT-Base      & 0.88 & 0.88 & 0.88 & 0.88 & 0.76 & 0.51 \\ 
\midrule 
 &  TM-mBERT           & 0.95 & 0.95 & 0.95 & 0.95 & 0.90 & 0.59 \\ 
&BanglaBERT &           0.85 & 0.87 & 0.86 & 0.84 & 0.71 & 0.43 \\ 
Approach 3 & mBERT-Base              & 0.92 & 0.93 & 0.92 & 0.92 & 0.85 & 0.48 \\ 
& DB-mBERT           & 0.94 & 0.94 & 0.94 & 0.94 & 0.89 & 0.54 \\ 
& BanglaBERT-Base      & \textbf{0.96} & 0.96 & 0.96 & \textbf{0.96} & 0.71 & 0.43 \\ 
\midrule
&TM-mBERT           & 0.90 & 0.90 & 0.90 & 0.90 & 0.81 & 0.57 \\ 
&BanglaBERT &           0.91 & 0.91 & 0.91 & 0.91 & 0.83 & 0.49 \\ 
Approach 4 & mBERT-Base   & 0.90 & 0.90 & 0.90 & 0.90 & 0.80 & 0.50 \\ 
& DB-mBERT           & 0.89 & 0.89 & 0.89 & 0.89 & 0.79 & 0.56 \\ 
& BanglaBERT-Base      & 0.93 & 0.93 & 0.93 & 0.93 & 0.86 & 0.48 \\ 
\botrule
\end{tabular}
\end{table}

\noindent\fbox{%
    \parbox{\textwidth}{%
        \textbf{RQ2: What impact does introducing summarization before fine-tuning have on fake news detection?}
    }%
}\\

We introduced the utilization of summarization technique in \textit{Approach 2}. Analyzing the results presented in Table \ref{restab: Test Dataset-1 test} and \ref{restab: Test Dataset-3 test}, we can observe that the BanglaBERT Base model achieved the highest accuracy scores of 88\% and 78\% on `Test Dataset-1' and `Test Dataset-3' respectively. The DB-mBERT model exhibited good performance on `Test Dataset-1' with an accuracy of 88\%, but its performance on `Test Dataset-3' was comparatively weaker, with an accuracy of only 70\%. When compared to the results of \textit{Approach 1}, we observed a slight decrease in performance after implementing the summarization technique in \textit{Approach 2}. However, this decrease is negligible, with only a marginal 1-2\% reduction in accuracy.\\

\noindent\fbox{%
    \parbox{\textwidth}{%
        \textbf{RQ3: How does the utilization of augmented fake data during fine-tuning affect fake news detection?}
    }%
}\\

In the third approach, we incorporated augmented fake news samples during the fine-tuning process and evaluated all the fine-tuned models using our three test datasets. It is important to note that `Test Dataset-2' was exclusively created for testing approaches involving augmentation, so it was not used to evaluate \textit{Approach 1} and \textit{Approach 2}. We only tested models in \textit{Approach 3} and \textit{Approach 4} where for fine-tuning we used the training dataset `Dataset 2'. As we have not used any translated news articles during the training in these approaches, this test corpus is completely unknown for the models \textit{Approach 3} and \textit{Approach 4}.  We show the performance comparison among the approaches on `Test Dataset-2' in Table \ref{restab: Test Dataset-2 test}. Considering the results on all three test datasets, the mBERT Base model emerged as the top performer in this approach. It achieved an accuracy score of 92\% on both `Test Dataset-1' as well as `Test Dataset-2' and 86\% on `Test Dataset-3'. The BanglaBERT Base model also demonstrated strong performance, achieving accuracy scores of 96\%, 93\% and 80\% on `Test Dataset-1', `Test Dataset-2' and `Test Dataset-3' respectively. We observed that the utilization of augmented fake data significantly improved the performance of all the fine-tuned models compared to approaches 1 and 2. While all models exhibited good performance, the mBERT Base and BanglaBERT Base models outperformed the others in terms of accuracy.\\

\begin{table}[h]
\centering
\caption{Performance comparison between Approach 3 and 4 on the second test dataset `Test Dataset-2'. \textit{Inference} represents performing classification without any training or fine-tuning.}\label{restab: Test Dataset-2 test}
\begin{tabular}{@{}llllllll@{}}
\toprule
    \textbf{Method}&\textbf{Model} & \textbf{A}    & \textbf{P}    & \textbf{R}    & \(\mathbf{F}_1\)   & \textbf{MCC}   & \textbf{ROC}   \\
\midrule     
    & TM-mBERT           & 0.37 & 0.22 & 0.37 & 0.28 & -0.37 & 0.61 \\ 
    & BanglaBERT & 0.50 & 0.25 & 0.50 & 0.33 & 0.00  & 0.50 \\
    Inference & mBERT-Base     & 0.51 & 0.51 & 0.51 & 0.45 & 0.02  & 0.41 \\ 
    & DB-mBERT           & 0.34 & 0.21 & 0.34 & 0.26 & -0.43 & 0.56 \\ 
    &  BanglaBERT-Base      & 0.48 & 0.41 & 0.48 & 0.35 & -0.09 & 0.50 \\ 
\midrule 
    & TM-mBERT           & 0.89 & 0.90 & 0.89 & 0.89 & 0.80 & 0.63 \\ 
    & BanglaBERT & 0.86 & 0.88 & 0.86 & 0.86 & 0.75 & 0.41 \\ 
    Approach 3 & mBERT-Base        & 0.92 & 0.93 & 0.92 & 0.92 & 0.85 & 0.48 \\ 
    & DB-mBERT           & 0.89 & 0.90 & 0.89 & 0.89 & 0.79 & 0.56 \\ 
    &  BanglaBERT-Base      & 0.93 & 0.93 & 0.93 & 0.93 & 0.87 & 0.49 \\
\midrule
    & TM-mBERT           & 0.91 & 0.91 & 0.91 & 0.91 & 0.82 & 0.51 \\ 
    & BanglaBERT & \textbf{0.97} & 0.98 & 0.97 & \textbf{0.97} & 0.95 & 0.53 \\ 
    Approach 4 & mBERT-Base   & 0.95 & 0.95 & 0.95 & 0.95 & 0.50 & 0.90 \\ 
    & DB-mBERT           & 0.91 & 0.91 & 0.91 & 0.91 & 0.82 & 0.52 \\ 
    &  BanglaBERT-Base      & 0.93 & 0.94 & 0.93 & 0.93 & 0.87 & 0.47 \\

\botrule
\end{tabular}
\end{table}

\noindent\fbox{%
    \parbox{\textwidth}{%
        \textbf{RQ4: What is the influence of summarizing augmented news articles on fake news classification?}
    }%
}\\

In the fourth approach, we introduced the technique of summarizing the augmented news articles. In this case, the BanglaBERT model exhibited exceptional performance across all three test datasets. It achieved accuracy scores of 91\%, 97\% and 82\% on `Test Dataset-1', `Test Dataset-2', and `Test Dataset-3' respectively. This model demonstrated superior learning capabilities when trained on summarized data compared to the other four models. The BanglaBERT-Base model also performed well, achieving an accuracy of 93\% on both `Test Dataset-1' as well as `Test Dataset-2' and 80\% on `Test Dataset-3'. Through empirical observation, we discovered that not all models adapt equally well to summarized training data. However, the BanglaBERT model showcased the highest performance when trained on summarized augmented data. When comparing these results with the previous approaches, it is evident that the summarization of augmented data positively influences the task of fake news classification.

\begin{table}[h]
\centering
\caption{Performance comparison across all the approaches on the third dataset `Test Dataset-3'. \textit{Inference} represents performing classification without any training or fine-tuning.}\label{restab: Test Dataset-3 test}
\begin{tabular}{@{}llllllll@{}}
\toprule
    \textbf{Method}&\textbf{Model} & \textbf{A}    & \textbf{P}    & \textbf{R}    & \(\mathbf{F}_1\)   & \textbf{MCC}   & \textbf{ROC}   \\
\midrule     
   & TM-mBERT        & 0.37 & 0.21 & 0.37 & 0.27 & -0.39 & 0.89 \\ 
& BanglaBERT & 0.50 & 0.25 & 0.50 & 0.33 & 0.00  & 0.50 \\ 
Inference & mBERT-Base     & 0.48 & 0.46 & 0.48 & 0.40 & -0.06 & 0.46 \\ 
& DB-mBERT           & 0.38 & 0.22 & 0.38 & 0.27 & -0.37 & 0.88 \\ 
&  BanglaBERT-Base      & 0.47 & 0.37 & 0.47 & 0.34 & -0.13 & 0.54 \\ 

\midrule
& TM-mBERT           & 0.73 & 0.82 & 0.73 & 0.70 & 0.54 & 0.73 \\ 
& BanglaBERT & 0.76 & 0.83 & 0.76 & 0.75 & 0.59 & 0.66 \\ 
Approach 1 & mBERT-Base    & 0.71 & 0.82 & 0.71 & 0.68 & 0.52 & 0.77 \\ 
&DB-mBERT            & 0.70 & 0.79 & 0.70 & 0.68 & 0.48 & 0.79 \\ 
& BanglaBERT-Base       & 0.80 & 0.84 & 0.80 & 0.79 & 0.64 & 0.63 \\

\midrule
&TM-mBERT           & 0.68 & 0.79 & 0.68 & 0.64 & 0.45 & 0.75 \\ 
&BanglaBERT & 0.73 & 0.82 & 0.73 & 0.71 & 0.54 & 0.83 \\ 
Approach 2  & mBERT-Base    & 0.74 & 0.81 & 0.74 & 0.72 & 0.54 & 0.75 \\ 
&DB-mBERT           & 0.70 & 0.79 & 0.70 & 0.67 & 0.48 & 0.80 \\ 
& BanglaBERT-Base      & 0.78 & 0.83 & 0.78 & 0.77 & 0.60 & 0.84 \\ 
\midrule 
&TM-mBERT           & 0.79 & 0.85 & 0.79 & 0.78 & 0.64 & 0.73 \\ 
&BanglaBERT &         0.71 & 0.71 & 0.71 & 0.71 & 0.41 & 0.52 \\ 
Approach 3 & mBERT-Base  & \textbf{0.86} & 0.88 & 0.86 & \textbf{0.86} & 0.74 & 0.83 \\ 
&DB-mBERT           & 0.78 & 0.85 & 0.79 & 0.78 & 0.64 & 0.73 \\ 
& BanglaBERT-Base      & 0.80 & 0.83 & 0.80 & 0.79 & 0.63 & 0.81 \\
\midrule
&TM-mBERT           & 0.74 & 0.83 & 0.74 & 0.72 & 0.56 & 0.83 \\ 
&BanglaBERT &         0.82 & 0.86 & 0.82 & 0.81 & 0.68 & 0.75 \\ 
Approach 4 & mBERT-Base   & 0.75 & 0.80 & 0.75 & 0.75 & 0.55 & 0.78 \\ 
&DB-mBERT           & 0.74 & 0.80 & 0.74 & 0.74 & 0.54 & 0.73 \\ 
& BanglaBERT-Base      & 0.80 & 0.85 & 0.80 & 0.80 & 0.65 & 0.70 \\ 

\botrule
\end{tabular}
\end{table}

\subsection{Result Analysis}
We analyze the results in two ways - quantitatively and qualitatively.\\

\begin{figure}[h]
    \centering
     \begin{subfigure}[b]{0.49\textwidth}
        \centering
        \includegraphics[width=\textwidth]{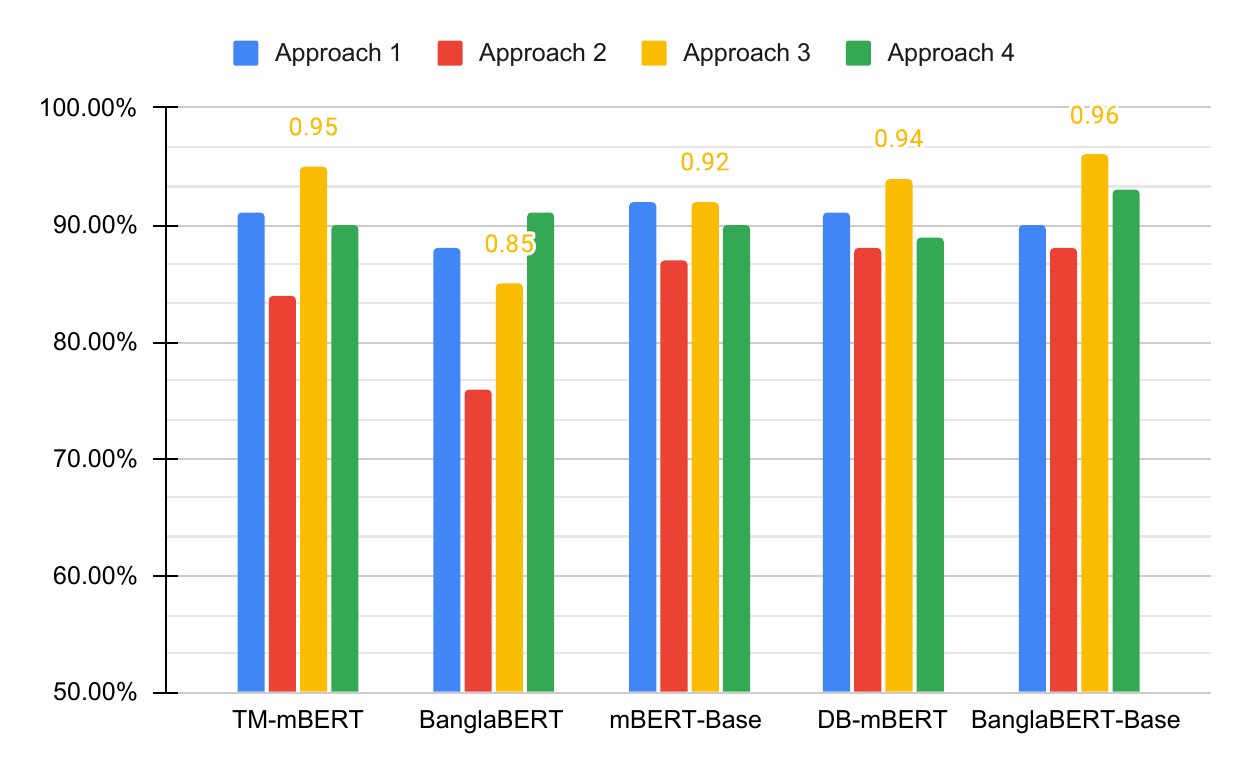}
        \caption{Comparison in terms of accuracy.} \label{test1-acc-chart}
    \end{subfigure}
    \hfill    
    \begin{subfigure}[b]{0.49\textwidth}
        \centering
        \includegraphics[width=\textwidth]{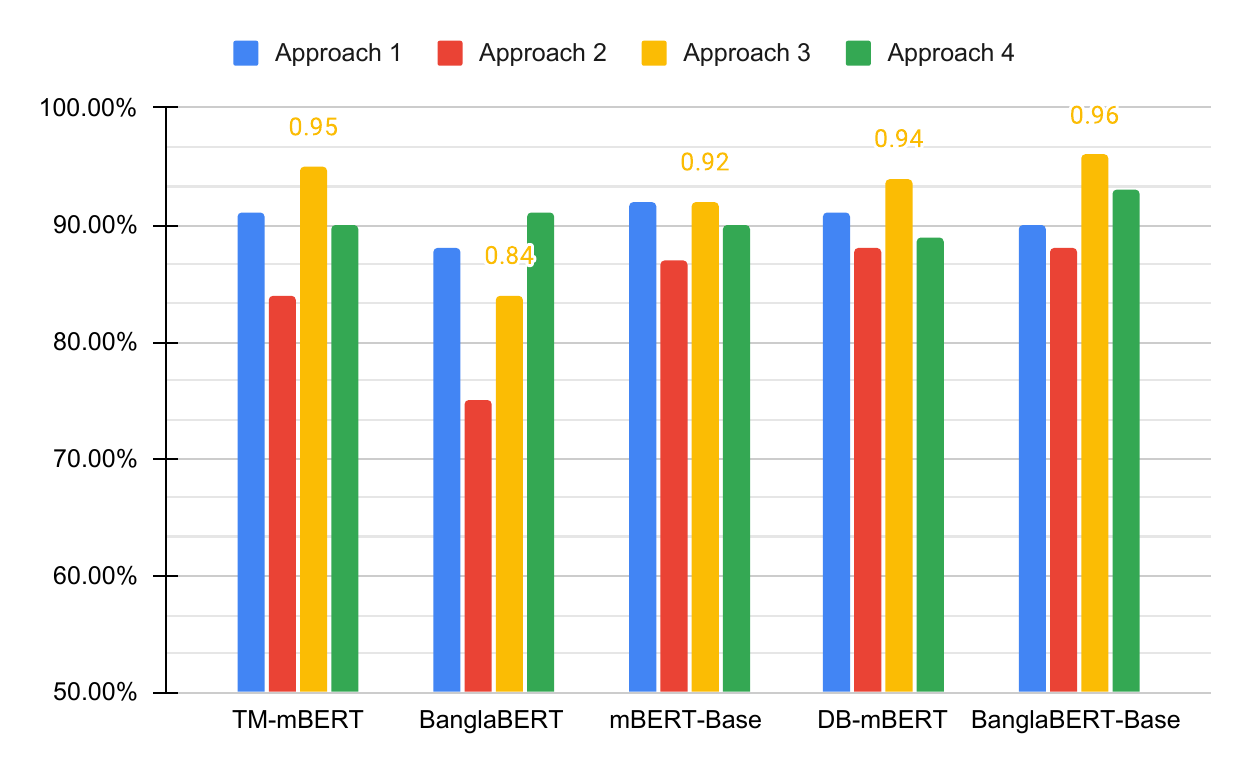} 
        \caption{Comparison in terms of \( F_1 \)-score.} \label{test1-F1-chart}
    \end{subfigure}        
\caption{Visual representation of the performance comparison on `Test Dataset-1'.}\label{fig:Test Dataset-1}
\end{figure}

\begin{figure}[h]
    \centering
     \begin{subfigure}[b]{0.49\textwidth}
        \centering
        \includegraphics[width=\textwidth]{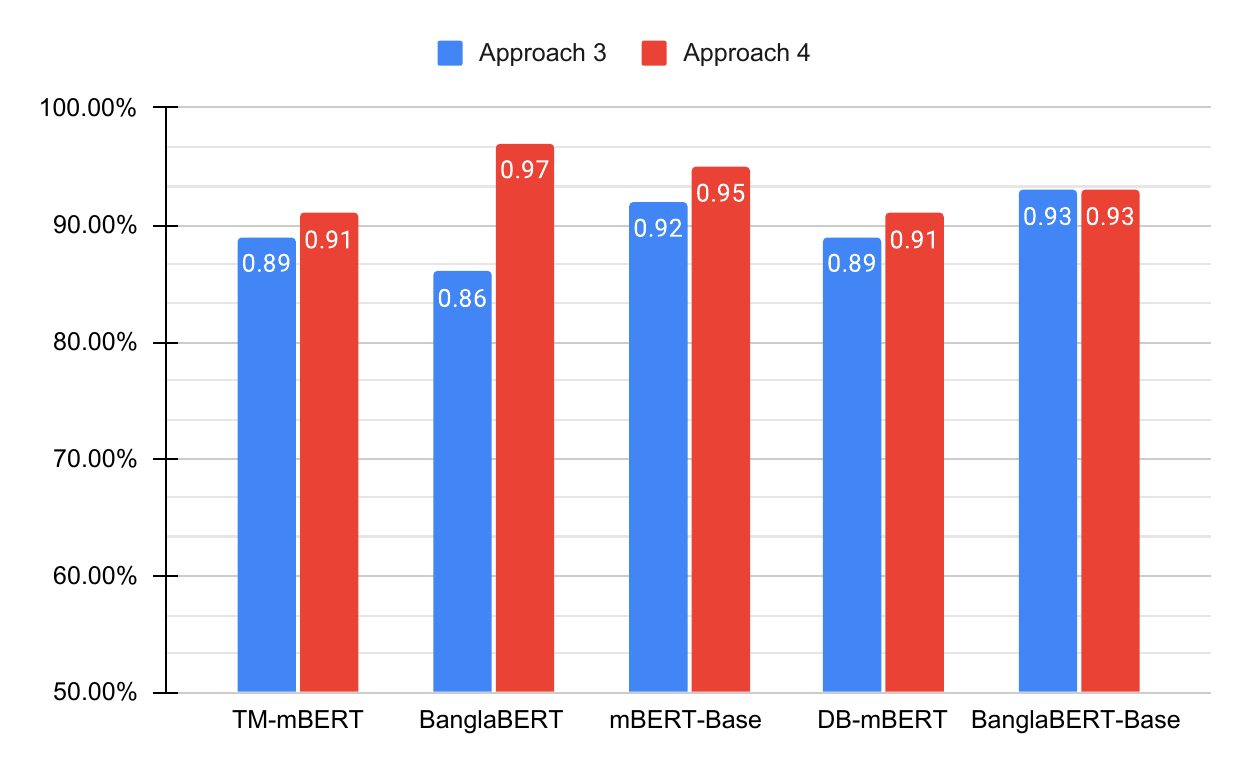}
        \caption{Comparison in terms of accuracy.} \label{test2-acc-chart}
    \end{subfigure}
    \hfill    
    \begin{subfigure}[b]{0.49\textwidth}
        \centering
        \includegraphics[width=\textwidth]{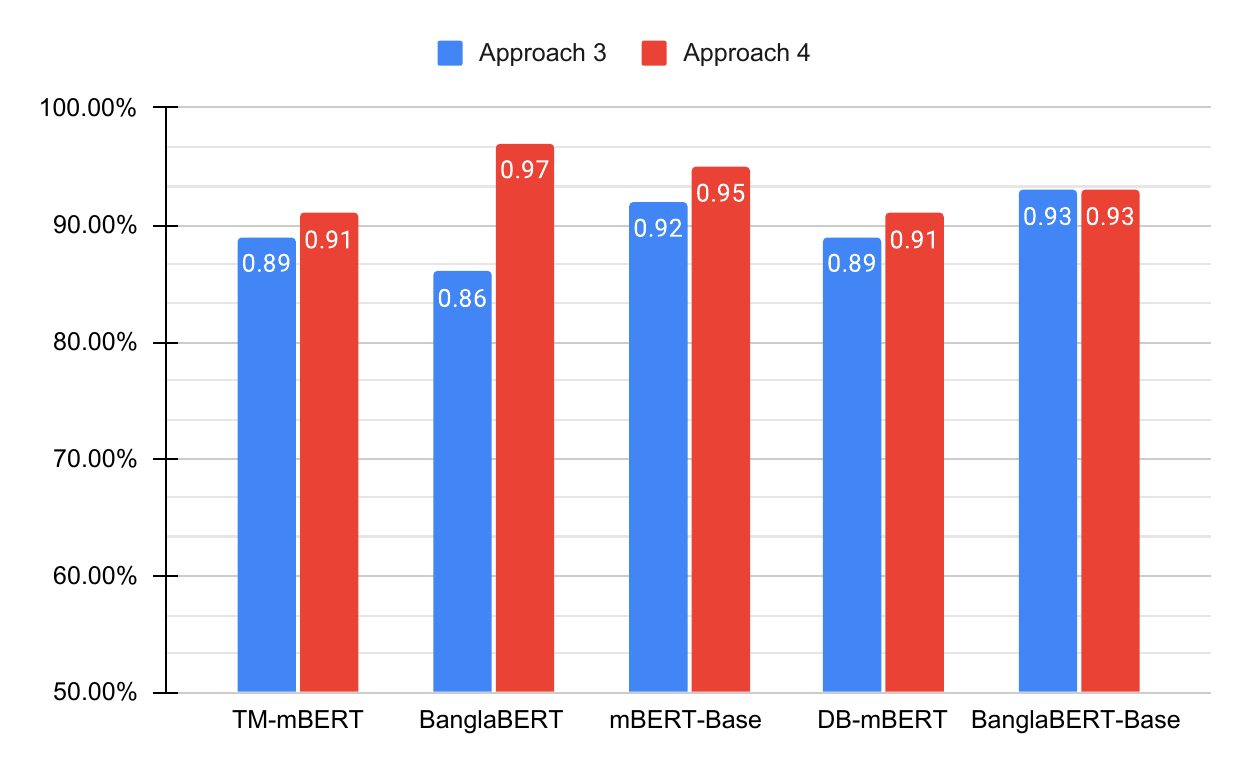} 
        \caption{Comparison in terms of \( F_1 \)-score.} \label{test2-F1-chart}
    \end{subfigure}        
\caption{Visual representation of the performance comparison on `Test Dataset-2'.}\label{fig:Test Dataset-2}
\end{figure}

\noindent\textbf{(a) Quantitative Analysis}\label{sec: quant-analysis}\\
This section seeks to experimentally support the models' performance. Among the three test datasets, we use the first two test datasets - `Test Dataset-1' and `Test Dataset-2' for quantitative analysis. Table \ref{restab: Test Dataset-1 test} and \ref{restab: Test Dataset-2 test} present the experimental findings that followed the evaluation of each model on these two test datasets. Figure \ref{fig:Test Dataset-1} is the visual representation of the performance comparison among all the fine-tuned models across all the approaches on `Test Dataset-1'.
As \textit{Approach 3} showed promising results on `Test Dataset-1', we decided to conduct further testing on this approach. In order to achieve a fair assessment, we excluded any samples from translated news articles during the training of models in \textit{Approach 3}. For this purpose, we created Test Dataset-2', ensuring that the translated data remained unknown to the models in Approach 3. This step allowed us to obtain a more comprehensive understanding of their true capability in detecting fake news. The performance comparison on `Test Dataset-2' is illustrated in Figure \ref{fig:Test Dataset-2}.

\begin{table}[!ht]
    \centering
    \caption{Average Text Length (in terms of word counts) for our Datasets.}
    \label{new_tab-1}
    \begin{tabular}{@{}lcc@{}}
    \toprule
        \textbf{Dataset name} & \textbf{Average Text Length} & \textbf{Longest Text Length} \\ 
    \midrule
        Dataset 1 & 313.9232228 & 5967 \\ \midrule
        Dataset 2 & 223.4071856 & 3030 \\ \midrule
        Test Dataset 1 & 279.9275 & 2712 \\ \midrule
        Test Dataset 2 & 311.4675 & 3275 \\ \midrule
        Test Dataset 3 & 242.1666667 & 1407 \\
    \botrule
    \end{tabular}
\end{table}

The results of our experiments demonstrate that both \textit{Approach 3} and \textit{Approach 4} perform exceptionally well in detecting fake news. \textit{Approach 4} achieved an impressive accuracy score of 97\%. It is worth noting that the incorporation of summarization improves the results for all models in \textit{Approach 4}, except for the BanglaBERT Base model which maintains the same level of performance. In \textit{Approach 1}, where `Dataset-1' was used for fine-tuning, we observed that summarization does not significantly enhance the performance in most cases. \textit{Approach 1} outperforms \textit{Approach 2} on `Test Dataset-1' and on `Test Dataset-3'. TM-mBERT, BanglaBERT, and BanglaBERT Base models performed worse in \textit{Approach 2}. However, DB-mBERT and mBERT Base models either perform equally or better. Comparing \textit{Approach 1} and \textit{Approach 3}, we find that \textit{Approach 3} generally performs better, except for the BanglaBERT model, which performs better in \textit{Approach 1} on both `Test Dataset-1' and `Test Dataset-3'. On the other hand, \textit{Approach 4} performed almost equally or better than \textit{Approach 1} on both `Test Dataset-1' and `Test Dataset-3'. Notably, \textit{Approach 3} exhibits significantly better performance than \textit{Approach 2}. When considering the impact of summarization, we observe that it decreases the performance in \textit{Approach 4} compared to \textit{Approach 3} on `Test Dataset-1' and `Test Dataset-3'. However, the BanglaBERT model is an exception to this trend, as it performs better after summarization on both of these test datasets. We conducted further testing with `Test Dataset-2', where summarization consistently improves performance. In \textit{Approach 4}, all models perform either equally or better than \textit{Approach 3} on `Test Dataset-2'. Based on the discussions, we conclude that all approaches exhibit good accuracy in detecting fake news. \textit{Approach 3} which incorporates augmentation, appears to perform better than the other approaches, but \textit{Approach 4} with the combination of augmentation and summarization, shows promising results.\\

\begin{figure}[h]
    \centering
     \begin{subfigure}[b]{0.49\textwidth}
        \centering
        \includegraphics[width=\textwidth]{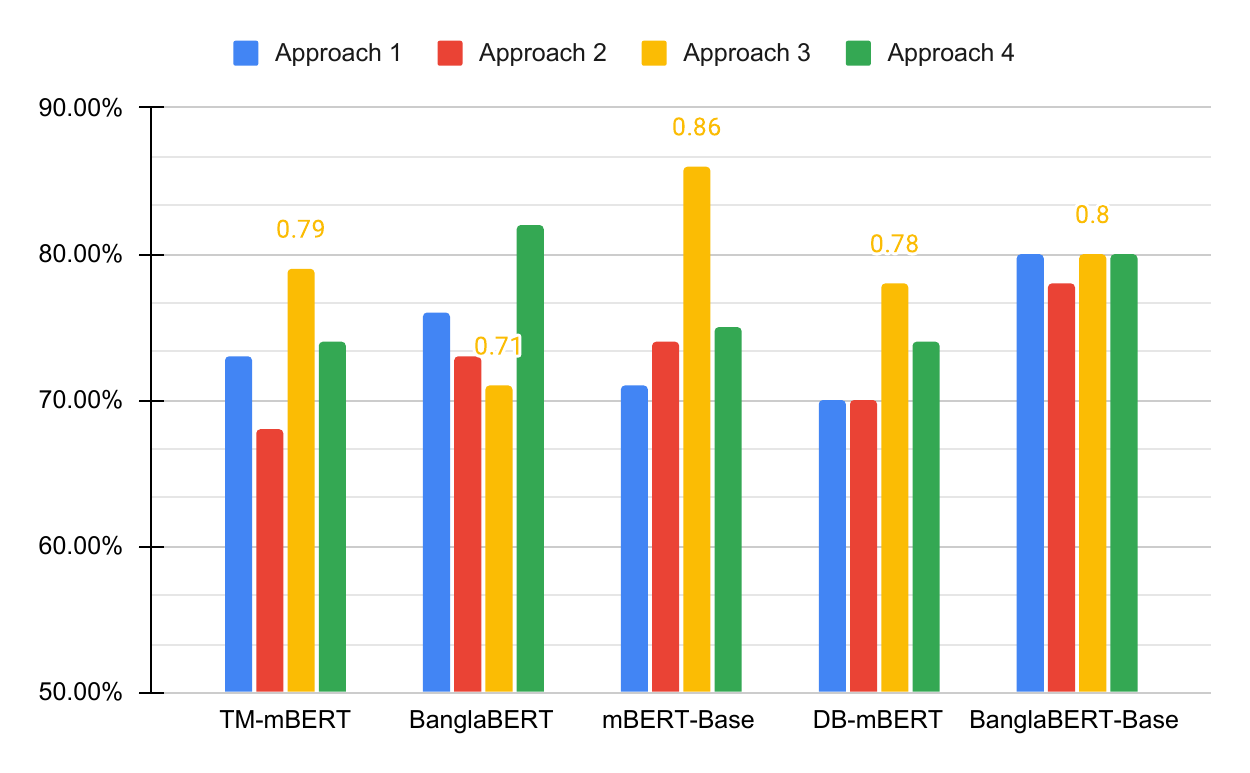}
        \caption{Comparison in terms of accuracy.} \label{test3-acc-chart}
    \end{subfigure}
    \hfill    
    \begin{subfigure}[b]{0.49\textwidth}
        \centering
        \includegraphics[width=\textwidth]{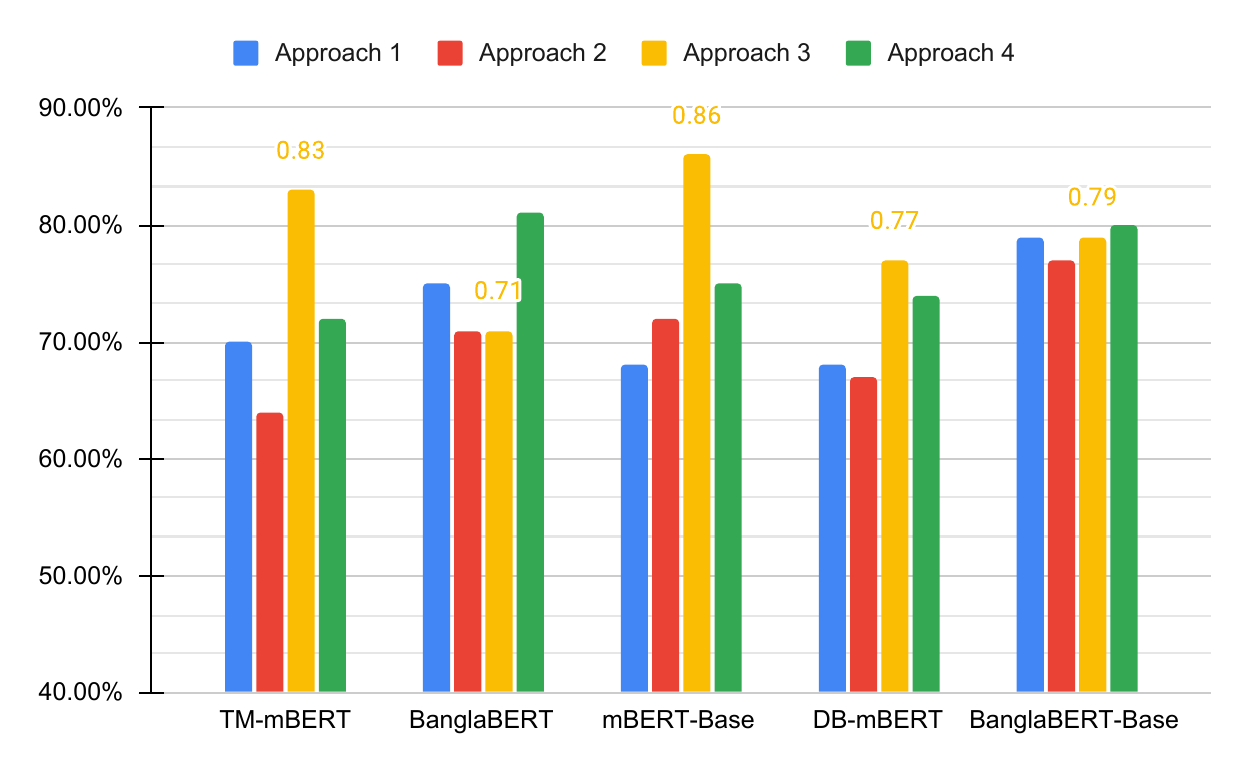} 
        \caption{Comparison in terms of \( F_1 \)-score.} \label{test3-F1-chart}
    \end{subfigure}        
\caption{Visual representation of the performance comparison on `Test Dataset-3'.}\label{fig:Test Dataset-3}
\end{figure}

\noindent\textbf{(b) Qualitative Analysis}\label{sec: qualit-analysis}\\
In this section, we evaluate the generalization ability of the model towards real-world test samples using `Test Dataset-3'. The results of the experiments on Test Dataset-3 are presented in Table \ref{restab: Test Dataset-3 test}. Even on this completely unseen test dataset, \textit{Approach 3} outperforms the other three approaches and achieves the highest accuracy and \( F_1 \) score of 0.86. The equality of the accuracy and \( F_1 \) score indicates that the model is capable of correctly classifying both positive and negative classes. Figure \ref{fig:Test Dataset-3} provides a visual representation of the performance comparison of all models on `Test Dataset-3'.
From the figure, we observe that the five transformer models we utilized exhibit varying performance characteristics. The TM-mBERT transformer model shows better performance without the use of summarization. It performs better in \textit{Approach 3} compared to \textit{Approach 1} indicating the effectiveness of the augmented training dataset. The BanglaBERT model performs exceptionally well in \textit{Approach 4}, where we employed summarization on the augmented dataset. The mBERT Base model excels in \textit{Approach 3} where the utilization of summarization is applied. However, we observe that summarization leads to a decrease in performance for this transformer model. Similarly, summarization adversely affects the performance of the DB-mBERT model. This model performs best in \textit{Approach 3}, but its performance decreases when summarization is implemented. On the other hand, BanglaBERT Base model displays a well-balanced performance across different strategies. It performs best in \textit{Approach 3} but it also achieves satisfactory scores in the other approaches.\\
\begin{table}[h]
    \centering
    \begin{tabular}{@{}lll@{}}
    \toprule
         \textbf{Models} & \textbf{Accuracy} & \(\mathbf{F}_1\)\textbf{-score}  \\
    \toprule
        TM-mBERT (inference) & 0.37 & 0.27 \\ 
        DB-mBERT (inference)  & 0.38 & 0.27 \\
         BanglaBERT Base (Approach 1) & 0.80 & 0.79 \\
         BanglaBERT Base (Approach 2) & 0.78 & 0.77 \\
        mBERT Base (Approach 3) & \textbf{0.86} & \textbf{0.86} \\
        BanglaBERT (Approach 4) & 0.82 & 0.81 \\
    \botrule
    \end{tabular}
\caption{Performance comparison of our approaches with existing pre-trained models on the third test dataset `Test Dataset-3'.}
\label{tab:comparison}
\end{table}

\noindent\textbf{(c) Summary of Result Analysis}\\
For Test Dataset-1 and Test Dataset-3, we observed that augmentation (Approach 3) is more effective than summarization only (Approach 2) as well as augmentation then summarization (Approach 4). However, in Table \ref{restab: Test Dataset-2 test}, we demonstrated that Approach 4 is more effective for Test Dataset-2. Our hypothesis is that Approach 4 performs better on longer news articles. We present the average text length (in terms of word count) for each dataset in Table \ref{new_tab-1}. The average text length for Test Dataset-2 is around 312 words. We speculate that in longer news articles, crucial information distinguishing fake from genuine news may reside towards the latter part. The pretrained transformer models utilized in this research can only process the initial 512 tokens of a text. It is conceivable that not all essential words in a lengthy news piece are encompassed within these 512 tokens. Therefore, with Approach 4, following augmentation and utilizing our summarization pipeline, there is an increased likelihood of encapsulating all vital information from a long news piece into a condensed form. This condensed text can then be tokenized into 512 tokens, encapsulating the most significant features of the news and aiding in the differentiation between fake and genuine news articles. Almost all models exhibited improvement, notably, the BanglaBERT model displayed an improvement in performance of nearly 11\%.

\begin{table}[!ht]
    \centering
    \caption{Experimental Results of different traditional machine learning and deep learning models on the `Test Dataset-3'.}
	\label{new_tab-3}
    \begin{tabular}{@{}lcccccccc@{}}
    
\toprule
\textbf{Model Name} & \multicolumn{2}{c}{\textbf{Approach 1}} & \multicolumn{2}{c}{\textbf{Approach 2}} & \multicolumn{2}{c}{\textbf{Approach 3}} & \multicolumn{2}{c}{\textbf{Approach 4}}  \\
& \textbf{Acc} & \(\mathbf{F}_1\) & \textbf{Acc} & \(\mathbf{F}_1\) & \textbf{Acc} & \(\mathbf{F}_1\)  & \textbf{Acc} & \(\mathbf{F}_1\)  \\
    
    \midrule
          SVC & 0.79 & 0.78 & 0.72 & 0.7 & 0.79 & 0.79 & 0.73 & 0.71 \\ 
        KNN & 0.69 & 0.66 & 0.73 & 0.72 & 0.75 & 0.75 & 0.71 & 0.7 \\ 
        DT & 0.72 & 0.7 & 0.63 & 0.61 & 0.72 & 0.7 & 0.66 & 0.64 \\ 
        RF & 0.67 & 0.64 & 0.78 & 0.78 & 0.73 & 0.71 & 0.68 & 0.64 \\ 
        ADB & 0.75 & 0.73 & 0.75 & 0.75 & 0.72 & 0.71 & 0.73 & 0.72 \\ 
        MNB & 0.75 & 0.74 & 0.79 & 0.79 & 0.81 & 0.81 & 0.78 & 0.77 \\ 
        LR & 0.79 & 0.79 & 0.7 & 0.67 & 0.79 & 0.78 & 0.75 & 0.73 \\ 
        bi-LSTM & 0.76 & 0.8 & 0.73 & 0.77 & 0.78 & 0.82 & 0.79 & 0.82 \\ 
        LSTM & 0.8 & 0.83 & \textbf{0.82} & 0.84 & 0.79 & 0.82 & \textbf{0.79} & 0.8 \\ 
        conv-HAN & \textbf{0.84} & 0.86 & 0.76 & 0.8 & \textbf{0.82} & 0.84 & 0.75 & 0.8 \\ 
    \botrule

    \end{tabular}
\end{table}

\subsection{Misclassification Analysis}\label{sec: misclassification analysis}

We conducted a brief error analysis on our best-performing model (mBERT Base - Approach 3) using Test Dataset-3. Out of 204 instances, it misclassified one real news article and 28 fake news articles. Table \ref{tab:misclassification_examples} shows examples of misclassified instances.\\
\noindent\textbf{(a) False Negative:} In the single false negative case, the model misclassified a real news article as fake, assigning it a very low confidence score (0.0047), indicating high uncertainty in its prediction. Our analysis revealed the following contributing factors:

\begin{enumerate}
    \item \textbf{Repetition of certain word: }The repeated use of the word “{\bng guNJ/jn}”\textit{[rumor]} may have misled the model into associating the article with speculation and misinformation.
    \item \textbf{Political Context: }The article discussed upcoming elections and political figures, which may have triggered features commonly associated with fabricated political narratives.
    \item \textbf{Entertainment Context: }Since the article involved a popular actor, the presence of entertainment-related content could have contributed to the model’s confusion, as entertainment articles are often exaggerated or fictionalized.
\end{enumerate}

\noindent\textbf{(b) False Positive:} Out of 28 false positive cases—fake news instances that were incorrectly predicted as real—we identified several common characteristics:

\begin{enumerate}
    \item \textbf{Numerical evidences: }Articles containing precise statistics, monetary values, or measurements (e.g., "21 crore USD") often appeared more credible to the model. This reliance on quantitative data as a proxy for truth led to misclassification.
    \item \textbf{Long texts: }Several misclassified fake articles were lengthy and detailed, which possibly increased their perceived credibility. The model may have associated longer texts with more thorough reporting, thus mistakenly classifying them as real.
    \item \textbf{Other sources as evidence: }Some fake articles attempted to build authenticity by mentioning other sources, organizations, or timelines. This inclusion of evidence-like elements misled the model, as it lacked the ability to verify these claims.
    \item \textbf{Lack of external reasoning: }BERT-based models are inherently limited in understanding real-world plausibility. As a result, even highly implausible or factually inaccurate articles were misclassified if their surface features aligned with real news patterns. For instance, stories about supernatural events or implausible medical conditions (e.g., a child growing animal fur) were predicted as real simply due to formal language and structured storytelling.
\end{enumerate}

\begin{table}[htbp]
\centering
\begin{tabular}{|p{8cm}|c|c|}
\hline
\textbf{News (Translated)} & \textbf{True Label} & \textbf{Predicted Label} \\
\hline
{\bng 'pRdhanmn/tRii ca{I}el AbshY{I} inr/bacn krb. baNNGlaedsh clic/cetRr jnipRy nayk ephredousek Ebar edkha eJet paer jatiiy sNNGsd inr/bacen, Emn guNJ/jn eshana Jaic/chl imiDya parhay. sm/pRit itin pRdhanmn/tRii eshkh Haisnar seNG/g raSh/TRiiy spher ghuer Eesechn. Aar Erpr ethek{I} E{I} guNJ/jenr shuru. Ebar E{I} guNJ/jenr jbab ideln ephredous. edesh ipher{I} inr/bacn pRseNG/g EkiT sNNGbad madhYem mukh khuleln E{I} Aibhenta. ephredous beln, Ekhena ikchu pha{I}nal Hyin. Emn ktha Aaim{O} shunich. sbar{I} menabasna thaek. Aamar{O} Aaech. jaygaTa Ekhn{O} {O}{I}rkm Abs/thaet{I} Aaech. Ekhn{O} ikchu{I} Hyin. Aamaek Jid na Aphar ker Aaim HNNYa ba na bleba ekmn ker. Aamaek Ekhn{O} dl ethek Aphar kra Hyin. Aphar na krel ejar ker ikchu kra Jaeb na. dl ethek pRs/tab ed{O}ya Hel AbshY{I} inr/bacn krb. itin Aar{O} beln, Aaim ik bleba buejh UThet parich na. Aaim pRdhanmn/tRiir seNG/g spher igeyich. ETa Aamar Aenk brh pa{O}ya. Erpr Uin Jid Aamaek iney ikchu ebheb thaekn taHel AalHamduillLaH. Aaim isenmar manuSh. kaejr madhYem manuShek ibenadn ideyich. Ekhn Aamar dWara Jid edsh {O} deshr Upkar Hy taHel AbshY{I} Aaim edsh {O} deshr jnY kaj kret ca{I}. manueShr esbay Jid inejek ineyaijt kret pair Er ecey esoubhaegYr Aar ikchu en{I}. }
[E] "If the Prime Minister wants, I will definitely run in the election." Rumors were circulating in the media that popular Bangladeshi film actor Ferdous might contest in the upcoming national elections. Recently, he accompanied Prime Minister Sheikh Hasina on a state visit, which fueled the speculation. Upon returning, Ferdous addressed the rumors in an interview, saying, "Nothing has been finalized yet. I’ve heard the rumors too. Everyone has their aspirations, and so do I. The situation is still in flux. If I am not offered a candidacy, how can I say yes or no? I haven’t been offered anything by the party yet. If they do, I will definitely run." He added, "I don't know what to say. Traveling with the Prime Minister was a great honor. If she sees potential in me, Alhamdulillah. I am from the film industry and have entertained people through my work. If I can serve the nation, there’s no greater honor." & 1 & 0 \\
\hline
{\bng cNNapa{I}nbabgNJ/j ejlay tasiphJa namk saerh itn bchr bJsii ishshur gaey pshur elam UThech. cNNapa{I}nbabgeNJ/j ibrl eraeg AakRan/t dirdR pirbaer saerh itn bcherr ishshu knYa tasiphJa jaHan muinra. tar baba idnmjur masuduj/jaman mamun. bairh naecal bnibbhaegr paesh egaDaUn paDa.ishshu tasiphJa jenMr pr ethek{I} tar shriierr lmWa lmWa pshm edkha Jay. idn Jt{I} grhaec/ch pshmguil{O} barhet barhet pshur meta edkha Jaec/ch. ishshu tasiphJar shriierr ipeThr echaT/T EkiT iTUmar ethek EiTr Ut//pit/t bel tasiphJar ma tanijla khatun janan. }
[E] In Chapainawabganj district, a three-and-a-half-year-old girl named Tasfia is growing "animal-like" hair on her body. She suffers from a rare condition. Her father, Masuduzzaman Mamun, is a day laborer, and they live in Godown Para, near the Nachol Forest Department. Since birth, long hair has been growing on Tasfia's body. As time passes, the hair is growing thicker, giving her an animal-like appearance. According to her mother, Tanzila Khatun, the condition appears to originate from a small tumor on her back. & 0 & 1 \\
\hline
\end{tabular}
\caption{Misclassification analysis example. Label 0 denotes fake and 1 denotes real.}
\label{tab:misclassification_examples}
\end{table}

\subsection{Comparison with Existing Models}\label{sec: comparison}
To evaluate and compare the performance of our proposed approaches, we conducted direct inference on two existing pre-trained models: TM-mBERT and DB-mBERT which are specifically designed for detecting Bengali fake news. Table \ref{tab:comparison} shows result comparison. We use `Test Dataset-3' for performance comparison between existing models and our approaches, because this test dataset is unknown to all the models. All our approaches performed better than the existing pre-trained models. 

We also conducted experiments using seven traditional machine learning models, including Support Vector Classifier (SVC), KNN, DT, RF, AB, MNB, and LR. Furthermore, we explored three deep learning techniques: Long Short Term Memory network (LSTM), Bidirectional Long Short Term Memory network (Bi-LSTM), and Convolutional Hierarchical Attention Network (Conv HAN). We trained these models using the `Dataset 2' training dataset, maintaining a 90:10 training-to-validation split and testing was conducted using the `Test Dataset-3'. Default hyperparameters were employed during training for all models. The experimental results are summarized in Table \ref{new_tab-3}.

For traditional machine learning models, textual data were numerically represented using Term Frequency-Inverse Document Frequency (TF-IDF). For deep learning approaches, we first preprocessed text using TensorFlow Keras' Tokenizer with a 10,000-word vocabulary, converting sequences to fixed-length 100-token representations through post-padding. All neural models (LSTM, BiLSTM, and Conv-HAN) employed 128-dimensional trainable embeddings. The Conv-HAN architecture specifically integrated 1D convolutions (64 filters, kernel size=3), max-pooling, an LSTM layer, and hierarchical attention mechanisms. Each deep learning model utilized a sigmoid output layer and was trained for 5 epochs (batch size=32) using Adam optimization, with consistent hyperparameters across architectures.

According to the findings presented in Table \ref{new_tab-3}, traditional machine learning models, except RF, ADB, and LR, demonstrated their best performance when augmentation was applied before training (Approach 3). RF and ADB attained the highest accuracy scores through summarization before training (Approach 2). Conversely, LR exhibited optimal performance without summarization or augmentation (Approach 1). Among the deep learning models, LSTM achieved the highest accuracy of 81.8\% when summarization was employed before training (Approach 2), while Bi-LSTM attained 78.9\% accuracy by employing both augmentation and summarization before training (Approach 4). However, akin to RF, Conv-HAN achieved its highest accuracy without any augmentation or summarization (Approach 1). Nonetheless, none of the traditional or deep learning models surpassed the performance of our fine-tuned mBERT-Base on `Test Dataset-3', as mBERT-Base achieved an accuracy of 86\% using augmentation (Approach 3).


\section{Conclusion and Future Work}\label{sec7}
In this paper, we presented our approach to classify Bengali fake news using a combination of summarization and augmentation techniques with pre-trained language models. To ensure a thorough evaluation of our trained models, we conducted tests on three distinct test datasets. The results revealed that our models achieved remarkably high levels of accuracy and \( F_1 \)-score on the first two test datasets. For the third test dataset, which was kept separate to assess generalization, the best model demonstrated an accuracy and \( F_1 \)-score of 86\%. These findings demonstrate the effectiveness of the model in accurately distinguishing between fake and authentic news articles. There are several promising directions for future work that can expand upon our research and broaden the scope of this paper. While our primary focus was on the classification of Bengali fake news, our approach holds potential for application in other languages with limited resources. Furthermore, our research primarily centered around the binary classification of authentic and fake news articles. To further advance the field, future studies could explore the realm of multi-class classification, involving the categorization of news articles into different types of fake news, including satire, propaganda, and clickbait.
Additionally, future work could explore the use of generative models for data augmentation, such as BigBird, which can handle longer documents and capture document-level semantics. Moreover, evaluating newer large language models (LLMs), which continue to evolve rapidly in terms of performance and adaptability, could lead to more robust and context-aware fake news classification systems.

\section*{Statements and Declarations}
\subsection{Ethical Approval and Consent to participate}
Not applicable.
\subsection{Consent for publication}
Not applicable.
\subsection{Human and Animal Ethics}
Not applicable.
\subsection{Availability of data}
The datasets generated during and/or analysed during the current study are available at - \url{https://github.com/arman-sakif/Bengali-Fake-News-Detection}
\subsection{Code availability}
The implementations can be found at - \url{https://github.com/arman-sakif/Bengali-Fake-News-Detection}
\subsection{Competing interests}
The authors have no competing interests to declare that are relevant to the content of this article.
\subsection{Funding}
No funding was received for conducting this study.
\bibliography{sn-bibliography}

\appendix
\section*{Appendix}

\section{CustomFake Corpus Creation}
\noindent\textbf{(a) Data Collection}. We visited the news websites listed by Hossain et al. \cite{bib10} as well as additional sources for our study. The names of these websites are listed in Table \ref{tab:sources}. In total, we gathered 102 new Bengali fake news articles. Additional details about news sources are provided in Table \ref{new_tab-2}. Given that we solely utilized this corpus to assess the generalization capabilities of our proposed methods, ensuring authenticity verification and manual annotation is essential. To carry out authenticity verification and manual annotation, we adhered to the same procedure outlined by Hossain et al. \cite{bib10}.\\
\noindent\textbf{(b)Annotation Process}. We chose three annotators from a group of seven potential candidates. Using a Google Form, we presented them with both headlines and articles. Initially, they were asked to determine whether the news was fake or authentic. Depending on their response to the first question, they were presented with a subsequent question offering four options for explaining why they believed the news to be fake or authentic. These four options were selected from those provided in \cite{bib10}.\\
\noindent\textbf{(c) Annotator Recruitment}. The three annotators participating in our study specialize in natural language processing tasks. Among them, two are final-year students from the Department of Mass Communication \& Journalism, while the other one is a faculty member at a respected educational institutions. We selected these individuals from a pool of seven candidates based on their trustworthiness scores, as determined through an evaluation process \cite{price2020six}. They were presented with 30 news articles and tasked with determining their authenticity. We randomly sampled 20 authentic news articles from the popular Bengali news portal ``Prothom Alo''\footnote{\url{https://www.prothomalo.com/}} and selected 10 fake news articles from the `BanFakeNews' dataset. Following completion of the task, we evaluated the number of correctly labeled samples for each candidate, ultimately selecting annotators who achieved a trustworthiness score above 85\%.\\
\noindent\textbf{(d) Annotation Quality}. In response to the first question, the \( F_1 \)-scores for determining the fake class among the three annotators are as follows: 65\%, 70\%, and 63\% respectively. The inter-annotator agreement, assessed using Fleiss' Kappa \cite{fleiss1971measuring} stands at 44.23\%, indicating that our human annotators provided identical responses in nearly 44\% of the cases. Regarding the second question, when the news is authentic, an average of 72\% of respondents select `The content is believable', while 20\% opt for `Source is reliable'. Here, the term `source' refers to an individual or organization capable of validating the reported news. Conversely, when the news is deemed fake, an average of 51\% of respondents indicate `The content is unrealistic', while 39\% suggest `Has no trustworthy source' as the reason.

\begin{center}

\begin{longtable}{>{\centering\hspace{0pt}}m{0.137\linewidth}>{\hspace{0pt}}m{0.269\linewidth}>{\hspace{0pt}}m{0.167\linewidth}>{\hspace{0pt}}m{0.36\linewidth}}
\caption{Details of the CustomFake dataset news sources (as of 14 May 2024).}
\label{new_tab-2}\\
\hline
\textbf{Article ID} & \multicolumn{1}{>{\centering\hspace{0pt}}m{0.269\linewidth}}{\textbf{Date}} & \multicolumn{1}{>{\centering\hspace{0pt}}m{0.167\linewidth}}{\textbf{Category}} & \multicolumn{1}{>{\centering\arraybackslash\hspace{0pt}}m{0.36\linewidth}}{\textbf{Source}}  \endfirsthead 
\hline
1                   & August 8, 2022                                                              & international                                                                   & Banladesh football ultra                                                                     \\
2                   & August 7, 2022                                                              & international                                                                   & The Daily star Bangla                                                                        \\
3                   & August 8, 2022                                                              & international                                                                   & Qatar Airways                                                                                \\
4                   & August 9, 2022                                                              & National                                                                        & The Daily star                                                                               \\
5                   & August 30, 2022                                                             & National                                                                        & Cinegolpo                                                                                    \\
6                   & August 31, 2022                                                             & international                                                                   & Cinegolpo                                                                                    \\
7                   & August 31, 2021                                                             & National                                                                        & The Daily star Bangla                                                                        \\
8                   & September 1, 2022                                                           & National                                                                        & The Daily star Bangla                                                                        \\
9                   & September 2, 2022                                                           & National                                                                        & Bangla Tribune                                                                               \\
10                  & September 3, 2022                                                           & international                                                                   & jachai.org                                                                                   \\
11                  & September 3, 2022                                                           & National                                                                        & songrami71                                                                                   \\
12                  & September 4, 2022                                                           & international                                                                   & sports protidin                                                                              \\
13                  & September 5, 2022                                                           & international                                                                   & Earki                                                                                        \\
14                  & September 6, 2022                                                           & international                                                                   & Earki                                                                                        \\
15                  & September 7, 2022                                                           & National                                                                        & Earki                                                                                        \\
16                  & September 8, 2022                                                           & National                                                                        & Earki                                                                                        \\
17                  & September 28, 2022                                                          & National                                                                        & jachai.org                                                                                   \\
18                  & September 29, 2022                                                          & international                                                                   & hindustantimes.com                                                                           \\
19                  & April 5, 2022                                                               & National                                                                        & jachai.org                                                                                   \\
20                  & October 1, 2022                                                             & National                                                                        & banglainsider.com                                                                            \\
21                  & September 11, 2022                                                          & National                                                                        & banglainsider.com                                                                            \\
22                  & September 8, 2018                                                           & National                                                                        & jamuna.tv                                                                                    \\
23                  & April 1, 2022                                                               & international                                                                   & jachai.org                                                                                   \\
24                  & March 7, 2022                                                               & international                                                                   & kalerkontho                                                                                  \\
25                  & April 8, 2022                                                               & international                                                                   & 71news24                                                                                     \\
26                  & November 19, 2022                                                           & national                                                                        & ntv                                                                                          \\
27                  & October 11, 2022                                                            & national                                                                        & kalerkontho                                                                                  \\
28                  & November 8, 2022                                                            & national                                                                        & jachai.org                                                                                   \\
29                  & April 16, 2022                                                              & international                                                                   & jachai.org                                                                                   \\
30                  & March 21, 2022                                                              & national                                                                        & Jugantor                                                                                     \\
31                  & January 16, 2022                                                            & international                                                                   & jachai.org                                                                                   \\
32                  & May 17, 2022                                                                & international                                                                   & jachai.org                                                                                   \\
33                  & May 18, 2022                                                                & international                                                                   & roarmedia                                                                                    \\
34                  & June 27, 2022                                                               & international                                                                   & zeenews                                                                                      \\
35                  & June 28, 2022                                                               & international                                                                   & zeenews                                                                                      \\
36                  & June 29, 2022                                                               & international                                                                   & zeenews                                                                                      \\
37                  & February 2, 2022                                                            & international                                                                   & Anandabazar                                                                                  \\
38                  & September 11, 2022                                                          & international                                                                   & awamiweb                                                                                     \\
39                  & September 12, 2022                                                          & international                                                                   & Bangladesh24Online                                                                           \\
40                  & May 22, 2022                                                                & international                                                                   & Bangladesh24Online                                                                           \\
41                  & June 21, 2021                                                               & international                                                                   & jagonews                                                                                     \\
42                  & February 6, 2021                                                            & international                                                                   & jagonews                                                                                     \\
43                  & September 14, 2018                                                          & international                                                                   & Aviation NewsBD                                                                              \\
44                  & May 23, 2017                                                                & international                                                                   & Kolar kontho                                                                                 \\
45                  & April 5, 2017                                                               & international                                                                   & jachai.org                                                                                   \\
46                  & April 3, 2017                                                               & international                                                                   & jachai.org                                                                                   \\
47                  & March 13, 2017                                                              & international                                                                   & jachai.org                                                                                   \\
48                  & July 11, 2017                                                               & national                                                                        & jachai.org                                                                                   \\
49                  & October 12, 2022                                                            & international                                                                   & hindustantimes.com                                                                           \\
50                  & August 31, 2022                                                             & international                                                                   & hindustantimes.com                                                                           \\
51                  & July 25, 2022                                                               & international                                                                   & hindustantimes.com                                                                           \\
52                  & November 14, 2022                                                           & international                                                                   & hindustantimes.com                                                                           \\
53                  & July 21, 2022                                                               & international                                                                   & hindustantimes.com                                                                           \\
54                  & November 16, 2022                                                           & national                                                                        & daily-star                                                                                   \\
55                  & June 16, 2021                                                               & national                                                                        & daily-star                                                                                   \\
56                  & February 4, 2022                                                            & national                                                                        & daily-star                                                                                   \\
57                  & May 9, 2022                                                                 & international                                                                   & newschecker.in                                                                               \\
58                  & November 1, 2022                                                            & international                                                                   & newschecker.in                                                                               \\
59                  & October 21, 2022                                                            & international                                                                   & newschecker.in                                                                               \\
60                  & November 19, 2022                                                           & national                                                                        & daily-star                                                                                   \\
61                  & May 30, 2018                                                                & national                                                                        & nationalistview.com                                                                          \\
62                  & March 3, 2020                                                               & national                                                                        & nationalistview.com                                                                          \\
63                  & December 29, 2019                                                           & national                                                                        & nationalistview.com                                                                          \\
64                  & May 19, 2021                                                                & national                                                                        & nationalistview.com                                                                          \\
65                  & April 10, 2021                                                              & national                                                                        & nationalistview.com                                                                          \\
66                  & August 28, 2021                                                             & national                                                                        & DailyNews96.com                                                                              \\
67                  & January 3, 2023                                                             & international                                                                   & DailyNews96.com                                                                              \\
68                  & January 2, 2023                                                             & international                                                                   & DailyNews96.com                                                                              \\
69                  & January 3, 2023                                                             & international                                                                   & bddailynews69.bd                                                                             \\
70                  & January 4, 2023                                                             & international                                                                   & priyobangla24.com                                                                            \\
71                  & May 21, 2022                                                                & national                                                                        & priyobangla24.com                                                                            \\
72                  & May 22, 2022                                                                & international                                                                   & bengali.news18.com                                                                           \\
73                  & May 23, 2022                                                                & international                                                                   & bengali.news18.com                                                                           \\
74                  & April 25, 2020                                                              & international                                                                   & bangla.hindustantimes.com                                                                    \\
75                  & December 31, 2022                                                           & international                                                                   & bangla.hindustantimes.com                                                                    \\
76                  & July 7, 2022                                                                & international                                                                   & bangla.hindustantimes.com                                                                    \\
77                  & January 5, 2023                                                             & national                                                                        & earki                                                                                        \\
78                  & December 26, 2022                                                           & national                                                                        & earki                                                                                        \\
79                  & December 27, 2022                                                           & national                                                                        & earki                                                                                        \\
80                  & December 28, 2022                                                           & national                                                                        & earki                                                                                        \\
81                  & December 25, 2022                                                           & national                                                                        & earki                                                                                        \\
82                  & April 30, 2022                                                              & national                                                                        & earki                                                                                        \\
83                  & July 15, 2022                                                               & national                                                                        & earki                                                                                        \\
84                  & January 1, 2023                                                             & national                                                                        & earki                                                                                        \\
85                  & January 3, 2023                                                             & international                                                                   & earki                                                                                        \\
86                  & January 3, 2023                                                             & international                                                                   & earki                                                                                        \\
87                  & January 5, 2023                                                             & international                                                                   & earki                                                                                        \\
88                  & May 26, 2019                                                                & international                                                                   & bengali.news18                                                                               \\
89                  & April 27, 2022                                                              & international                                                                   & bengali.news19                                                                               \\
90                  & November 23, 2022                                                           & international                                                                   & earki                                                                                        \\
91                  & January 3, 2023                                                             & national                                                                        & earki                                                                                        \\
92                  & January 2, 2023                                                             & international                                                                   & bddailynews69.bd                                                                             \\
93                  & January 5, 2023                                                             & international                                                                   & bddailynews69.bd                                                                             \\
94                  & January 5, 2023                                                             & international                                                                   & bddailynews69.bd                                                                             \\
95                  & January 6, 2023                                                             & international                                                                   & bddailynews69.bd                                                                             \\
96                  & January 10, 2023                                                            & international                                                                   & bddailynews69.bd                                                                             \\
97                  & January 11, 2023                                                            & national                                                                        & earki                                                                                        \\
98                  & January 12, 2023                                                            & national                                                                        & earki                                                                                        \\
99                  & January 11, 2023                                                            & national                                                                        & earki                                                                                        \\
100                 & January 9, 2023                                                             & national                                                                        & earki                                                                                        \\
101                 & January 8, 2023                                                             & national                                                                        & earki                                                                                        \\
102                 & December 25, 2022                                                           & international                                                                   & bengali.news19                                                                               \\
\hline
\end{longtable}

\end{center}

\end{document}